\documentclass[10pt,twocolumn,letterpaper]{article}

\usepackage[pagenumbers]{cvpr} 
\usepackage{amsmath}
\usepackage{makecell} 
\usepackage{subcaption}
\usepackage{booktabs, makecell, xcolor, array, colortbl}
\usepackage{booktabs}
\usepackage{siunitx} 
\usepackage{tikz}

\usepackage[most]{tcolorbox}   
\newcommand{\logoicon}{%
  \raisebox{-0.20\height}{\includegraphics[height=1.2em]{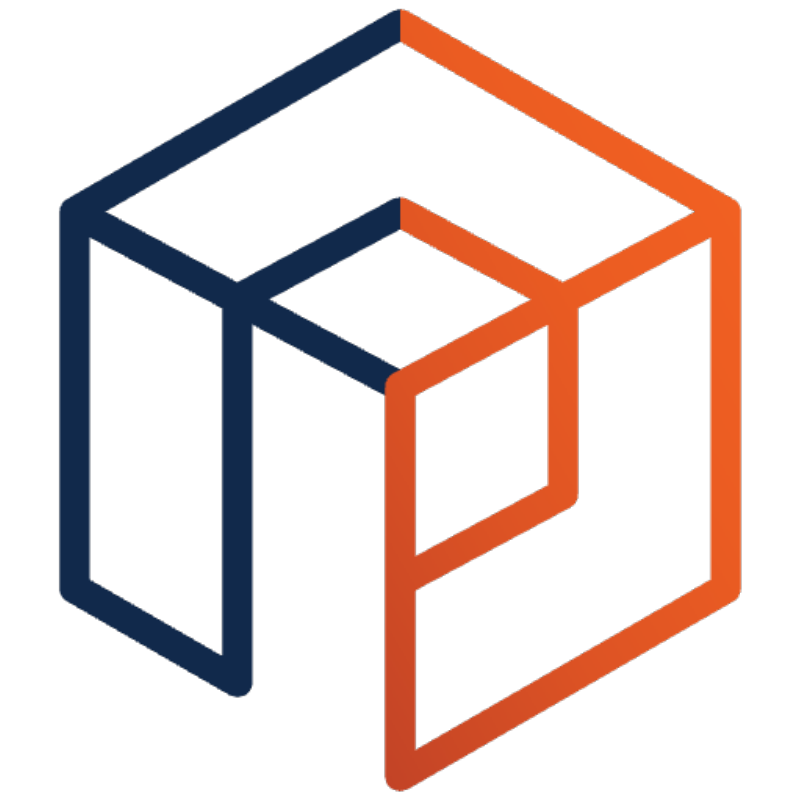}}%
}

\definecolor{promptbg}{RGB}{248,248,248}
\definecolor{promptframe}{RGB}{210,210,210}
\definecolor{promptnumbers}{RGB}{150,150,150}
\definecolor{promptkw}{RGB}{0,92,184}

\lstdefinestyle{promptstyle}{
  basicstyle=\ttfamily\footnotesize,
  backgroundcolor=\color{promptbg},
  frame=single,
  rulecolor=\color{promptframe},
  numbers=left,
  numberstyle=\tiny\color{promptnumbers},
  numbersep=6pt,
  showstringspaces=false,
  breaklines=true,
  breakatwhitespace=true,
  columns=fullflexible,
  xleftmargin=1.5em,
  xrightmargin=0.5em,
  aboveskip=0.7\baselineskip,
  belowskip=0.7\baselineskip,
  keywordstyle=\color{promptkw}\bfseries
}

\usepackage[T1]{fontenc}
\usepackage{microtype}           
\usepackage{nicefrac}            
\usepackage{xspace}              
\usepackage{amsmath, amssymb, amsfonts, amsthm, mathtools, mathrsfs,bm}

\usepackage{booktabs}            
\usepackage{nicematrix}         
\usepackage{multirow}
\usepackage{makecell}
\usepackage{bigstrut}
\usepackage{enumitem}            
\setitemize{label=\textbullet, leftmargin=*, nolistsep}
\usepackage{graphicx}
\usepackage{adjustbox}           
\usepackage{float}               
\usepackage{wrapfig}             
\usepackage{subcaption}          
\expandafter\def\csname ver@subfig.sty\endcsname{}  
\usepackage[table]{xcolor}
\usepackage{xcolor}
\usepackage{fontawesome5}        
\usepackage{pifont}              
\usepackage{tikz}
\usepackage{circledsteps}
\definecolor{cvprblue}{rgb}{0.21,0.49,0.74}
\usepackage[pagebackref,breaklinks,colorlinks,allcolors=cvprblue]{hyperref}
\usepackage[all]{hypcap}         
\usepackage{url}                 
\usepackage{algorithm}
\usepackage{algorithmic}
\usepackage[normalem]{ulem}      
\usepackage{rotating}            
\usepackage{stackengine}         
\usepackage{caption}             
\usepackage{listings}            
\usepackage{soul}                
\usepackage{gradient-text}       
\usepackage{pgfplots}            
\pgfplotsset{compat=newest}
\usepackage{pgfplotstable}       
\usepackage{tikz}                
\usepackage{svg}                 

\usepackage{cuted}
\usepackage{tcolorbox}
\newtcolorbox{planbox}[1]{
    colback=gray!5,
    colframe=gray!75,
    title=#1,
    fonttitle=\bfseries
}

\renewcommand{\ie}{\emph{i.e.},\xspace}
\renewcommand{\eg}{\emph{e.g.},\xspace}
\renewcommand{\etc}{\textit{etc}.\xspace}      

\newcommand{\cmark}{\textcolor{ForestGreen}{\ding{51}}}  
\newcommand{\xmark}{\textcolor{red}{\ding{55}}}     

\def\vg{{\bm{g}}}

\def\vp{{\bm{p}}}

\def\vx{{\bm{x}}}

\def\vz{{\bm{z}}}
\def\vI{{\bm{I}}}
\def\vM{{\bm{M}}}

\newcommand{\Enc}{\operatorname{Enc}}
\newcommand{\Dec}{\operatorname{Dec}}
\newcommand{\FMDenoise}{\operatorname{FMDenoise}}

\newcommand{\softmax}{\operatorname{softmax}}

\newcommand{\Normal}{\mathcal N}

\newcommand{\tcircle}[1]{\tikz[baseline=(X.base)] \node[draw,circle,inner sep=1pt, line width=0.3pt](X){\scriptsize #1};}
\newcommand{\bftcircle}[1]{\tikz[baseline=(X.base)] \node[draw,circle,inner sep=1pt, line width=0.6pt](X){\scriptsize #1};}
\definecolor{IllinoisOrange}{HTML}{FF5F05}
\definecolor{IllinoisBlue}{HTML}{13294B}

\definecolor{rewardpurple}{RGB}{133,76,248}
\definecolor{flowgreen}{RGB}{170,219,102}

\newcommand{\modelnamecolor}{\textbf{\textcolor{rewardpurple}{Reward}\textcolor{flowgreen}{Flow}}}

\newcommand{\modelname}{%
  \textbf{Reward{Flow}}\xspace}

\newcommand{\modelnamenc}{RewardFlow\xspace}

\usepackage[percent]{overpic}

\definecolor{cvprblue}{rgb}{0.21,0.49,0.74}
\usepackage[pagebackref,breaklinks,colorlinks,allcolors=cvprblue]{hyperref}

\def\confName{CVPR}
\def\confYear{2026}

\title{\raisebox{-3pt}{\includegraphics[width=3cm]{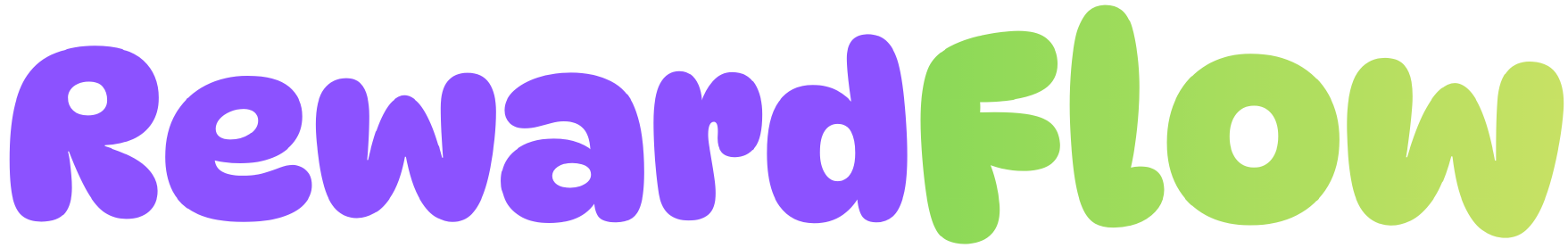}}: Generate Images by Optimizing What You Reward}

\author{
Onkar Susladkar$^{1}$ \hspace{0.6em}
Dong-Hwan Jang$^{1}$ \hspace{0.6em}
Tushar Prakash$^{2}$ \hspace{0.6em}
Adheesh Juvekar$^{1}$ \hspace{0.6em}
Vedant Shah$^{1}$ \\[0.3em]
Ayush Barik$^{1}$ \hspace{0.6em}
Nabeel Bashir$^{1}$ \hspace{0.6em}
Muntasir Wahed$^{1}$ \hspace{0.6em}
Ritish Shrirao$^{2}$ \hspace{0.6em}
Ismini Lourentzou$^{1}$ \\[0.5em]
$^{1}$University of Illinois Urbana-Champaign \qquad
$^{2}$Sony Research, India \\[0.4em]
\texttt{\{onkarks2, lourent2\}@illinois.edu}
}

\begin{document}

\twocolumn[{%
\renewcommand\twocolumn[1][]{#1}%
\maketitle
\vspace{-0.8cm}
\centering
\includegraphics[width=0.99\linewidth]{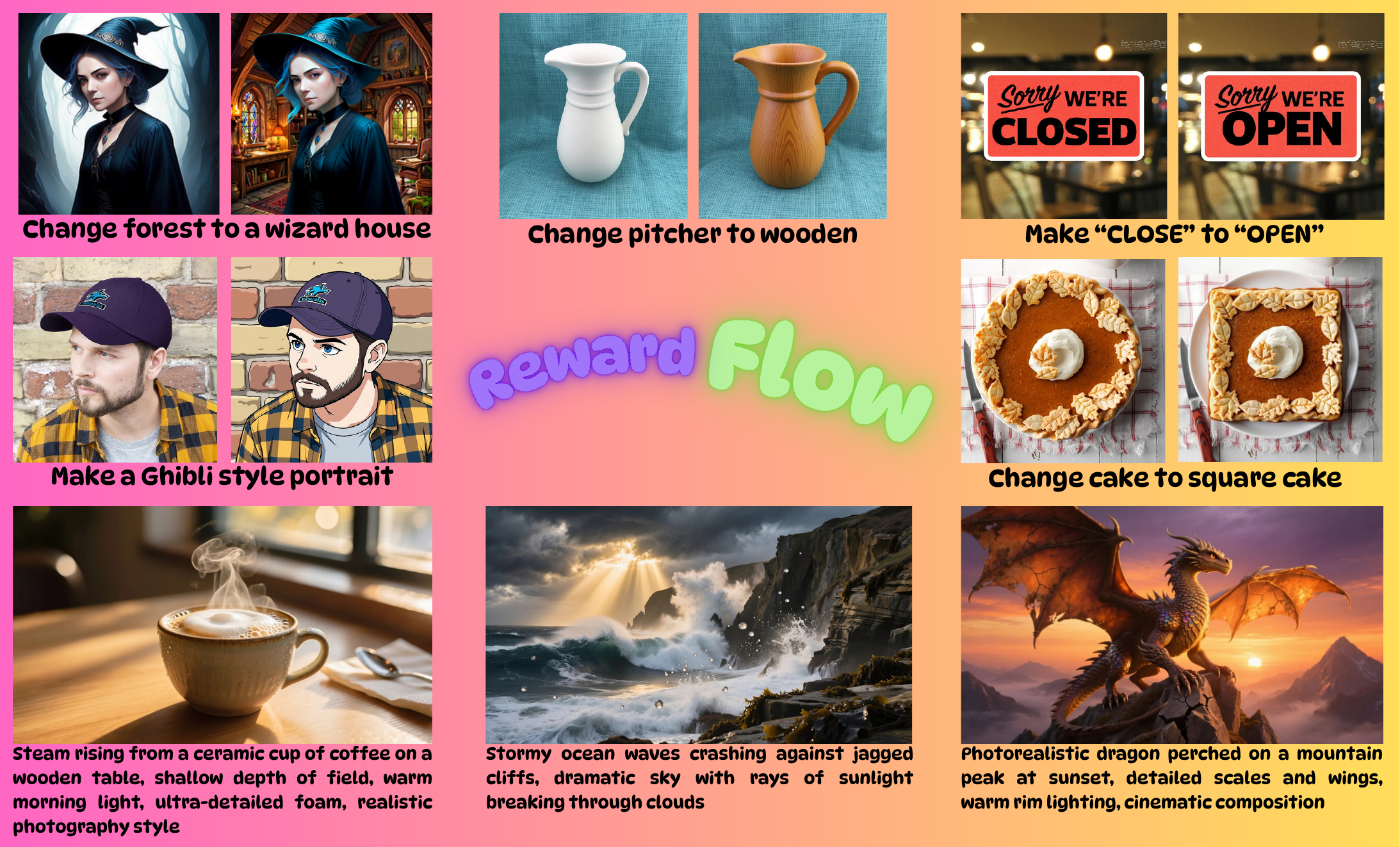}
\vspace{-0.3cm}
\captionof{figure}{\modelname{} enables accurate, localized, inversion-free image editing and generation using multi-reward Langevin guidance.}
\label{fig:teaser}
\vspace{0.5cm}
}]

\begin{abstract}
We introduce \modelnamenc, an inversion-free framework that steers pretrained diffusion and flow-matching models at inference time through multi-reward Langevin dynamics. \modelnamenc unifies complementary differentiable rewards for semantic alignment, perceptual fidelity, localized grounding, object consistency, and human preference, and further introduces a differentiable VQA-based reward that provides fine-grained semantic supervision through language-vision reasoning. To coordinate these heterogeneous objectives, we design a prompt-aware adaptive policy that extracts semantic primitives from the instruction, infers edit intent, and dynamically modulates reward weights and step sizes throughout sampling. 
Across several image editing and compositional generation benchmarks, \modelnamenc delivers state-of-the-art edit fidelity and compositional alignment.

\noindent \logoicon~\href{https://plan-lab.github.io/rewardflow}{\textcolor{IllinoisBlue}{PLAN Lab}~\textcolor{IllinoisOrange}{https://plan-lab.github.io/rewardflow}}
\end{abstract}    
\section{Introduction}

Text-guided image generation and editing have become the most active frontiers in generative modeling, driven by recent advances in diffusion and flow-matching models~\cite{ho2020denoising,song2020score,lipman2022flow,blackforest2025flux,esser2024scaling}. The ability to generate or modify an image based solely on natural-language instructions has enabled diverse applications in visual design, content creation, and interactive editing. While significant progress has been made in fine-tuning–based approaches ~\cite{ruiz2023dreambooth,brooks2023instructpix2pix,lu2024regiondrag, gao2025lora}, these methods require expensive optimization and exhibit limited generalization beyond the training distribution. Consequently, training-free and inversion-free methods have emerged as a practical alternative. By operating directly on pretrained models at inference time without modifying model weights, such methods offer broad applicability and efficient deployment~\cite{gong2025instantedit, hertz2022prompt, ju2024directinv,kulikov2025flowedit, bai2024edicho, zhu2025training}.

However, on one hand, inversion often distorts layout or identity and introduces brittle forward–reverse sampling loops.
On the other hand, inversion-free methods circumvent reconstruction but lose access to a faithful latent representation of the original image. As a result, they frequently suffer from content drift, semantic leakage, weak object localization, and insufficient fine-grained controllability.
More recently, reward-guided frameworks~\cite{eyring2024reno, xie2025dymo, min2025origen} attempt to address controllability, but they employ coarse rewards with weak semantic grounding, lack adaptive policies, and provide no mechanism for harmonizing heterogeneous objectives over the sampling trajectory. These shortcomings prevent existing methods from achieving precise, consistent, and semantically faithful edits in a truly zero-shot setting.

To address these challenges, we propose \modelnamecolor, a zero-shot, training-free, and inversion-free framework for text-guided image editing and generation based on \textbf{multi-reward Langevin dynamics}. \modelnamenc fuses complementary signals, including global semantics, perceptual alignment, spatial grounding, aesthetic quality, and semantic faithfulness, into a differentiable objective that guides a pretrained flow-matching model at inference time. 
Specifically, we introduce two new reward formulations that significantly improve localized accuracy and compositional alignment: (i) a SAM2~\cite{ravi2024sam} text-guided object reward that produces differentiable localization signals, enforcing mask-consistent edits and penalizing leakage outside target regions, and (ii) a differentiable VQA reward that enforces fine-grained semantic correctness through language–vision reasoning. 

To balance coarse-to-fine scheduling of these heterogeneous reward signals, we introduce a novel \textbf{prompt-aware adaptive policy} that extracts semantic primitives from the instruction, and dynamically adjusts reward weights throughout the denoising process, enabling stable convergence and improving sampling efficiency. 
Furthermore, to prevent drift during inference-time optimization, we incorporate a \textbf{clean-latent KL tether} that anchors the sampling trajectory to the original latent representation. As shown in ~\Cref{fig:teaser}, \modelnamenc supports high-quality text-to-image generation and precise edits across diverse instruction types, including localized style, attributes, and text modifications. 

\noindent In summary, our contributions are:
\begin{itemize}
    \item We introduce \modelnamecolor{}, a training-free multi-reward-guided Langevin framework that integrates complementary differentiable signals to enable controllable, inversion-free editing and generation. Across multiple benchmarks, \modelnamenc achieves state-of-the-art zero-shot performance in editing fidelity and compositional generation.
    \item We design a novel \textbf{prompt-aware adaptive policy} that parses semantic primitives from the text instruction, infers intent, and dynamically modulates reward weights,  providing coarse-to-fine efficient optimization.
    \item We propose a novel \textbf{differentiable VQA-based reward} that provides fine-grained semantic supervision, ensuring accurate attribute changes and improved compositional alignment, alongside a \textbf{SAM-guided reward} that supports localized edits, penalizing leakage outside target regions.
    \item Moreover, we provide a principled theoretical justification, showing that our update corresponds to a valid discretization of a Langevin SDE targeting a prompt-tilted density, establishing a sound foundation for stable reward-guided convergence under our framework.
\end{itemize}

\section{Related Work}

\noindent \textbf{Training-Based Methods.}
Early image generation and editing approaches rely on fine-tuning large diffusion~\cite{blattmann2023stable} or GAN~\cite{goodfellow2020generative} models, often achieving high fidelity but at significant computational cost.
DreamBooth~\cite{ruiz2023dreambooth} fine-tunes text-to-image models on a few subject images to bind unique identities, while Imagic~\cite{kawar2023imagic} enables identity-preserving edits from a single image. Other approaches modify latent representations rather than model weights, \eg StyleCLIP~\cite{patashnik2021styleclip} manipulates StyleGAN~\cite{karras2019style} latents using CLIP~\cite{radford2021learning} guidance, and Textual Inversion~\cite{gal2022image} introduces new token embeddings without retraining. Although training-based methods exhibit strong alignment, they require model updates and do not generalize well to unseen edits, making them impractical for interactive applications.

\noindent \textbf{Inference-Time Controllable Generation and Editing.}
Diffusion inversion seeks a noise latent whose denoising trajectory exactly reconstructs the source image.
Recent methods such as Null-text inversion~\cite{mokady2023null}, Direct Inversion~\cite{ju2024directinv}, and LEDITS++~\cite{brack2024ledits++} achieve faithful reconstructions but require costly forward–reverse passes and rely heavily on accurate inversion.
A parallel line of work performs inference-time editing without training or inversion. SDEdit~\cite{meng2021sdedit} and PostEdit~\cite{tian2024postedit}
perturb and denoise the input image with stochastic or posterior sampling, while FlowEdit~\cite{kulikov2025flowedit}, FreeFine~\cite{zhu2025training}, and Edicho~\cite{bai2024edicho}
steer pre-trained models through ODE paths, attention control, or correspondence cues.
Despite their speed and convenience, these methods lack a faithful latent reconstruction of the input image, which often leads to content drift, weak identity and layout preservation, hallucinated details, and limited fine-grained controllability.

\noindent \textbf{Reward-Guided Optimization.}
Recent work explores reward-based alignment of generative models.
ReNO~\cite{eyring2024reno} optimizes latent trajectories based on multi-objective feedback, while ORIGEN~\cite{min2025origen} applies reward-guided Langevin sampling for zero-shot grounding. Recent formulations~\cite{chang2025training} cast this as trajectory optimal control.
While effective for global alignment, existing methods lack localized reward models and adaptive control, often leading to drift, over-editing, or weak spatial consistency.
Our work introduces a unified Langevin dynamics framework that integrates a suite of coarse-to-fine rewards and a prompt-aware policy to adaptively and precisely steer generation.

\section{\modelnamecolor{} Method} 
\label{sec:method}
\modelnamenc{} performs training-free, reward-guided generation 
by treating each denoising step as an instance of test-time optimization over a set of differentiable rewards. 
The key idea is to evaluate a set of rewards on the intermediate decoded image, combine them through a \textbf{prompt-aware adaptive policy} that dynamically sets time-varying reward weights and step sizes (\Cref{subsec:adaptive}), and map the resulting fused gradient back into the latent space through the decoder–denoiser chain rule. This produces a reward-guided drift term that augments the native flow-matching dynamics. 
Our method integrates a \textbf{diverse set of heterogeneous rewards}  (\Cref{subsec:rewards}), covering semantic, perceptual, regional, object-level, and QA-based, into a unified guidance signal that steers the sampling trajectory of a pretrained flow-matching model. An \textbf{identity-preserving KL tether} further anchors the generation to the source image to preserve identity and layout  (\Cref{subsec:kltether}).

\subsection{Multi-Reward Langevin-Based Generation}
\label{subsec:prelim-langevin}
Given a text prompt $\vp$ and optional input image $\vx$, we obtain the initial clean latent $\vz_0\!=\!\mathrm{Enc}(\vx)$ by encoding the image (for image-conditioned cases) or sampling from noise (for unconditional text-to-image generation). We then initialize the forward trajectory  $\vz^{(0)}\!=\!\alpha_{\bar t} \vz_0\!+\!\sigma_{\bar t}\varepsilon$, with $\varepsilon\!\sim\!\Normal(0,\vI)$ at a variance-preserving noise level $\bar t$ and begin reverse-time sampling. At each step $k$ with time $t_k$, a flow-matching (rectified-flow) denoiser produces a clean latent 
$\tilde \vz^{(k)}\!=\!\mathrm{Den}_\theta(\vz^{(k)},t_k,\vp)$ and its decoded image
$\vI^{(k)}\!=\!\mathrm{Dec}(\tilde \vz^{(k)})$. The decoded image $\vI^{(k)}$  is treated as an optimization variable evaluated by a set of differentiable rewards $\{R_i(\vI^{(k)},\vp)\}$. Each reward produces an image-space gradient $g_{\vI,i}^{(k)}\!=\!\nabla_{\vI^{(k)}} R_i(\vI^{(k)},\vp)$, which is mapped into the latent-space drifts using the decoder $J_{\mathrm{Dec}}(\cdot)$ and denoiser $J_{\mathrm{Den}}(\cdot)$ Jacobians, \ie
\begin{equation}
\label{eq:prelim-reward-drift}
\setlength{\abovedisplayskip}{6pt}
\setlength{\belowdisplayskip}{6pt}
g_{R_i,k}=\lambda_R\,
J_{\mathrm{Den}}(\vz^{(k)},t_k,\vp)^{\!\top}
J_{\mathrm{Dec}}(\tilde \vz^{(k)})^{\!\top}\, g_{\vI,i}^{(k)}.
\end{equation}

Because different rewards should dominate at different moments in the generation process, we fuse the individual reward signals into a total reward that is adaptively weighted based on (i) the text prompt, (ii) the current denoising time, and (iii) the evolving generation state. A lightweight policy (described in \Cref{subsec:adaptive}) predicts a time-dependent weight for each reward $w_i(t_k)$ and an adaptive step size $\eta_k$, which controls the magnitude of the update. 
We standardize each reward using its running mean $\mu_i$ and standard deviation $\sigma_i$, as 
$\bar R^{(k)}_i\!=\!(R^{(k)}_i-\mu_i)/(\sigma_i+\epsilon)$ to ensure consistent scaling. We then define the fused reward $R_{\mathrm{tot}}^{(k)}\!=\!\sum_i w_i(t_k)\,\bar R_i^{(k)}$, whose gradient is mapped to the latent space exactly as above, yielding the total reward drift $g_{R_{\mathrm{tot}},k}$. This mechanism naturally modulates the generation trajectory so that different objectives activate at appropriate times, for example, global semantic alignment early in the trajectory and fine spatial refinements later. The reverse-time update $t\!=\!\bar t \!\to\! 0$ is
\begin{equation}
\label{eq:update}
\setlength{\abovedisplayskip}{6pt}
\setlength{\belowdisplayskip}{6pt}
\begin{aligned}
\vz^{(k+1)} &= \vz^{(k)} + \eta_k\big(f_k + g_{R_{\mathrm{tot}},k} + g_{\mathrm{KL},k}\big) + \xi_k,\\
t_{k+1} &= t_k - \eta_k,\qquad
\xi_k \sim \mathcal N\!\big(0,\,2\,\gamma_{k}\,\eta_k\,\vI\big),
\end{aligned}
\end{equation}
where $f_k{=}v_\theta(\vz^{(k)},t_k,\vp)$ is the backbone drift,
$g_{R_{\mathrm{tot}},k}$ is the fused multi-head reward, and $g_{\mathrm{KL},k}$ is a clean-space KL tether (described in \Cref{subsec:kltether}) that preserves identity.

As we show in the supplementary material, this update corresponds to a valid discretization of a Langevin SDE that targets a prompt-tilted density, ensuring consistent convergence toward an image that satisfies the semantic, spatial, and structural constraints of the prompt.

\subsection{Prompt-Aware Adaptive Policy}
\label{subsec:adaptive}

A central contribution of our method is the adaptive policy that acts as a closed-loop controller for the Langevin sampler, and determines how strongly each reward should influence the trajectory and how aggressively the sampler should move at each step. 
Intuitively, different prompts require different forms of controllable generation, such as adding an object, removing a region, or changing a style attribute, each with distinct reward priorities and step sizes, and these requirements change over time as the image becomes closer to the target.
To capture this, rather than relying on fixed inference schedules, the policy \tcircle{1} extracts semantic primitives from $\vp$ once before sampling, and then each step dynamically adjusts \tcircle{2} reward weights $w_i$ and object direction $s_{obj}$, and \tcircle{3} the {reward-aware step size} ($\eta_k$).

\vspace{0.1cm}
\noindent \textbf{\bftcircle{1} Semantic Primitives (SP).}
As a one-time pre-process, before sampling, we parse the prompt $\vp$ once using an LLM to extract {Semantic Primitives} $\mathrm{SP}(\vp)$ as a set of atomic, actionable concepts parsed from $\vp$. Each primitive corresponds to a self-contained generative objective, \eg $\vp =$``Remove the cap from the person and add sunglasses'' $\rightarrow \mathrm{SP}(\vp)\!=\!\{$``Remove Cap'', ``add sunglasses''\}). SPs enables computing per-primitive perceptual, region, and object-level rewards, preventing interference between unrelated objectives and improving controllability.

\noindent \textbf{\bftcircle{2} Dynamic Reward Weighting.}
Different prompts require different reward strengths, and the optimal weighting changes over time as the generation progresses.
At each step, the policy therefore computes the reward weights $w_i$ using three sources of information.
First, for a given SP and the full prompt, we classify the generation intent into three coarse categories (\emph{add}, \emph{remove}, or \emph{style}) which broadly capture how content should evolve. These intent probabilities are
fused into a single {base profile} prior $\tau = \pi_{\text{add}}\tau^{(\text{add})} + \pi_{\text{remove}}\tau^{(\text{remove})} + \pi_{\text{style}}\tau^{(\text{style})}$,
where each reward template $\tau(\cdot)$ encodes the characteristic importance pattern of the differentiable rewards for that type of intent. For example, if the given prompt is removal-heavy, this profile downweights region/object rewards.
To avoid drift as sampling proceeds, we compute 
$\delta_i\!=\!\max(0, R_{i}^{(k-1)}\!-\!R_{i}^{(k)})$.

Moreover, we incorporate a lightweight schedule term $h_i(t_k)$ that nudges the policy to emphasize different reward families at different noise levels (\eg localization early, semantics late).
This prevents premature overfitting and stabilizes the trajectory.  
Because each prompt yields multiple SPs, weighting occurs in two stages. First, for each SP-level reward family, we compute $\tau_i$, $\delta_i$, and $h_i(t_k)$ for every SP-specific reward and apply a softmax across SPs to obtain one representative reward for that family.  Second, we combine these four representative SP-level rewards with the global prompt-level rewards using
$w_i(t_k)\!=\!\operatorname*{softmax}_i\!\big(\beta\,(\tau_i\!+\!\kappa_{\text{fb}}\delta_i\!+\!\kappa_{\text{sch}}h_i(t_k))\big)$, where $\kappa_{\text{fb}}$ and $\kappa_{\text{sch}}$, control the influence of the feedback and schedule terms, respectively, $\tau_i$ is the base profile prior, and $\beta$ is the softmax temperature.
\begin{figure}[t!]
\includegraphics[width=0.99\linewidth]{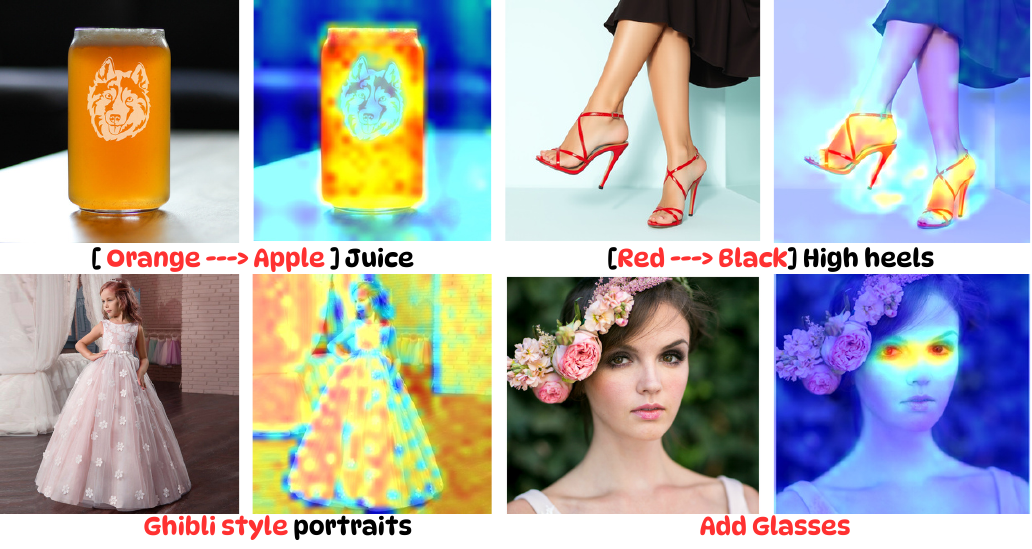}
    \vspace{-0.4cm}
    \caption{\textbf{Gradient localization of our differentiable rewards.}
    We visualize the image-space gradient $\nabla_I R_{tot}(\cdot)$ for various edit prompts.
    Our proposed rewards prevent \textit{semantic leakage} by concentrating the gradient precisely on target semantic regions, demonstrating the fine-grained spatial control enabled by \modelnamenc.}
    \label{fig:gradient_activation}
    \vspace{-0.4cm}
\end{figure}

\noindent\textbf{Object Direction.}
Object-level changes are inherently directional: some prompts require introducing an object while others require removing or suppressing an object. 
A single object reward cannot distinguish between these two cases on its own, unless we explicitly encode direction. To handle this, the policy predicts a direction multiplier based on the intent classifier 
$s_{obj}\!=\!\pi_{add}\!-\!\pi_{remove}$, and the final reward used per SP is $R_\text{obj}' = s_{obj} \cdot R_\text{obj}$.
A positive (add-intent) makes the sampler increase object presence ($s_{obj} \approx +1$), while a negative value (remove-intent) makes it decrease object presence ($s_{obj} \approx -1$).

\begin{table*}[t!]
\centering
\caption{\textbf{\textsc{PIE-Bench} image editing results.} 
\modelnamenc consistently improves edit fidelity and spatial localization across all metrics while maintaining competitive runtime.
Best results are in \textbf{bold} and strong baselines are \uline{underlined}.}
\vspace{-0.3cm}
\resizebox{0.97\linewidth}{!}{
\begin{tabular}{lcccccccccc}
\toprule
\textbf{Method} &
\makecell{\textbf{Distance} $\downarrow$ \\ ($\times10^3$)} &
\textbf{PSNR} $\uparrow$ &
\makecell{\textbf{LPIPS} $\downarrow$ \\ ($\times10^3$)} &
\makecell{\textbf{MSE} $\downarrow$ \\ ($\times10^4$)} &
\makecell{\textbf{SSIM} $\uparrow$ \\ ($\times10^2$)} &
\textbf{Whole} $\uparrow$ &
\textbf{Edited} $\uparrow$ &
\textbf{NFE} $\downarrow$ &
\textbf{Step} $\downarrow$ \\
\midrule
EF~\cite{huberman2024edit}     & 8.39 & 27.49 & 44.38 & 29.79 & 85.61 & 25.87 & 22.14  & 70  & 50 \\
ProxG~\cite{han2024proxedit}   & 8.39 & 28.45 & \textbf{38.27} & \textbf{25.63} & 85.87 & 25.04 & 21.64 & 100 & 50 \\
P2P~\cite{hertz2022prompt}     & 9.58 & 27.72 & 44.98 & 30.02 & 85.01 & 24.94 & 21.57  & 100 & 50 \\
DI~\cite{ju2024directinv}         & 11.60 & 27.25 & 49.25 & 32.87 & 84.86 & 25.83 & 22.39  & 100 & 50 \\
AREdit~\cite{wang2025training} & 13.12 & 28.78 & 42.67 & 37.88 & 87.19 & 26.57 & 24.51  & \uline{16}  &  30 \\
InfEdit~\cite{xu2023infedit}   & 13.87 & 28.63 & 39.80 & 33.19 & 86.28 & 25.84 & 22.44   & \textbf{12}  & 12 \\
TurboEdit~\cite{deutch2024turboedit} & 15.11  & 26.04  & 69.54 & 55.12 &  84.27& 26.09 & 23.35   & 24 & 12  \\
InstantEdit~\cite{gong2025instantedit} & 12.57 & \uline{29.63} & \uline{35.27} & \uline{24.57} & \uline{87.40} & 26.06 & 22.73   & 24  & {12} \\
FlowEdit~\cite{kulikov2025flowedit} & 11.56 & 28.33 & 43.57 & 37.48 & 86.23 & 26.43 & 23.03  & 33  & 30 \\
FlowChef~\cite{patel2025flowchef}   & 9.67  & 29.03 & 43.11 & 36.67 &  {87.44} & 27.05 & 23.09  & 28  & 30 \\
KV-Edit~\cite{zhu2025kvedit}   & \uline{8.47} & 29.04 & 43.44 & 35.47 & 86.66 & \uline{28.21} & \uline{23.92}  & 27 & 40 \\
\hline
\rowcolor{rewardpurple!10}
{Flux+\modelnamecolor}            & \textbf{7.78} & \textbf{31.21} & 40.55 & 26.47 & \textbf{89.67} & \textbf{29.44} & \textbf{26.62}  & 43 & 20 \\
\rowcolor{flowgreen!20}
{Qwen Image+\modelnamecolor}       & {7.64} & {32.09} & {38.47} & {23.57} & {90.21} & {29.78} & {27.57}  & 54 & 25 \\
\bottomrule
InfEdit~\cite{xu2023infedit}  & \uline{16.19} & 26.75 & 50.79 & 42.33 & 84.71 & 25.68 & 22.27  & \textbf{4} & {4} \\
TurboEdit~\cite{deutch2024turboedit} & 18.57 & 24.59 & 77.53 & 58.48 & 82.64 & 25.70 & 22.30   & \textbf{4} & {4} \\
InstantEdit~\cite{gong2025instantedit} & 17.14 & \uline{27.96} & \uline{44.39} & \uline{34.94} & \uline{86.44} & \uline{26.28} & \uline{22.82}   & \uline{8}   & 4 \\
\hline
\rowcolor{rewardpurple!10} {Flux+\modelnamecolor} & \textbf{13.44} & \textbf{29.57} & \textbf{40.03} & \textbf{29.66} & \textbf{88.13} & \textbf{27.31} & \textbf{24.69} & 6 & 4 \\
\rowcolor{flowgreen!20} {Qwen Image+\modelnamecolor} & {10.33} & {29.92} & {37.92} & {26.95} & {90.12} & {28.55} & {27.52}  & 6 & 4 \\
\bottomrule
\end{tabular}
}
\label{tab:PIE_bench}
\vspace{-0.3cm}
\end{table*}

\begin{figure*}
    \centering
    \includegraphics[width=.99\linewidth]{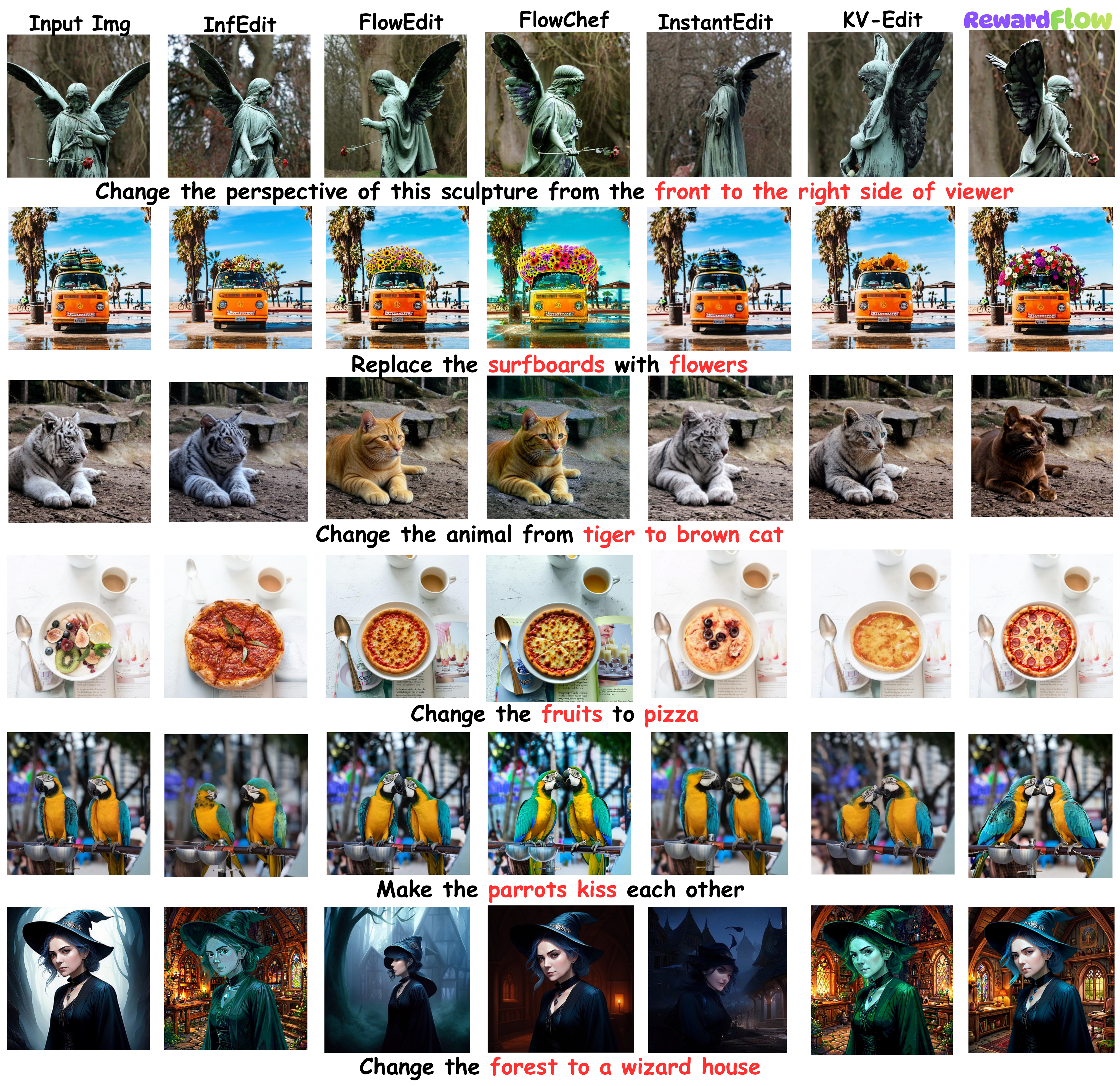}
    \vspace{-0.4cm}
    \caption{\textbf{Image editing qualitative comparison across diverse instruction types.} \modelnamenc produces edits that are both semantically accurate and spatially localized, while better preserving background structure, lighting, and identity compared to prior methods.}
    \label{fig:main_fig}
    \vspace{-0.4cm}
\end{figure*}

\noindent\textbf{\bftcircle{3} Reward-Aware Step Size ($\eta_k$).}
Finally, the policy adapts the step size $\eta_k$ based on the \textit{current} total reward $R_{\text{tot}}^{(k)}$.
A high reward indicates we are close to the target,
so we take smaller, more careful steps (refinement).
A low reward indicates we are far, prompting larger steps (exploration).
This is controlled by a logistic map
\begin{equation}
\setlength{\abovedisplayskip}{6pt}
\setlength{\belowdisplayskip}{6pt}
    \eta_k = \eta_{\min}+(\eta_{\max}-\eta_{\min})\cdot\sigma\!\big(-\gamma_\eta(R_{\text{tot}}^{(k)}-r_0)\big),
\end{equation}
where $\sigma(u)$ is the logistic function and $r_0$ target threshold.

\subsection{Differentiable Rewards}
\label{subsec:rewards}

Relying on a single, global reward offers semantic alignment but lacks spatial precision; its gradients tend to diffuse across the entire image, leading to semantic leakage where unrelated regions are unintentionally modified. To counter this, we construct a hierarchical reward toolkit that provides fine-grained spatial, perceptual, and object-level control.

We employ two levels of differentiable rewards: SP-level rewards, marked with $\circledast$
which are computed separately for each semantic primitive, and global prompt-level rewards, marked with $\circlearrowright$, which operate over the entire input.
\Cref{fig:gradient_activation} illustrates how these rewards focus gradients on the intended edit region, preventing spillover.

\noindent $\circledast$ \textbf{Global and Perceptual Alignment} ($R_{\text{glb}}$ and $R_{\text{per}}$).
To ensure overall semantic correctness, we compute two complementary alignment rewards for each semantic SP: the global alignment reward that measures cosine similarity between image and text embeddings from SigLIP~\cite{tschannen2025siglip} encoders and the perceptual alignment reward that applies the same cosine formulation using Perception encoders.

\noindent $\circledast$ \textbf{Region-level Grounding}  ($R_{\text{rg}}$).
Global alignment alone cannot ensure that the modified or generated content occurs in the correct image region. To provide spatial specificity, we compute region–text relevance scores between region proposals and the SP phrases using RegionCLIP-style embeddings~\cite{zhong2022regionclip}. These scores are then softly pooled using a temperature-controlled attention mechanism, yielding a differentiable reward that encourages content changes to appear in spatial areas that are most relevant to the prompt.

\noindent $\circledast$ \textbf{Object Consistency}  ($R_{\text{oc}}$).
Region grounding ensures localized generation in the right spatial region but does not guarantee that the correct object appears or disappears, or is modified as intended. 
To capture this object-level behavior, we employ text-conditioned SAM2~\cite{ravi2024sam} to obtain soft masks ${\vM_j}$ with confidence scores $a_j$ and mixture weights $\omega_j\!=\!\softmax(a_j/\tau_{\text{SP}})$ with $\tau_{\text{SP}}$ temperature.
For each SP, we compute an {object alignment score} $F_{\text{obj}}(\vM_j,\vg_{\text{SP}})$, which measures the cosine similarity between the masked image and the SP, minus a small “leakage” penalty for similarity outside the mask. 
The final reward 
$R_{\text{oc}}\!=\!\sum_j \omega_j F_{\text{obj}}(\vM_j,\vg_{\text{SP}})$ evaluates whether the intended object semantics are correctly realized and spatially confined to the appropriate region.

\noindent $\circlearrowright$ \textbf{Human Preference Alignment} ($R_{\text{hps}}$).
This reward is defined as the normalized scalar output of a differentiable predictor $H_{\text{HPS}}(\vI^{(k)}, \vp)$ (\eg HPS v2~\cite{wu2023hpsv2}) measuring image-prompt consistency: 
$R_{\text{hps}}\big(\vI^{(k)},\vp\big)\!=\!\mathrm{norm}\!\big(H_{\text{HPS}}(\vI^{(k)},\vp)\big)$.

\noindent $\circlearrowright$ \textbf{VQA Reward} ($R_{\text{vqa}}$).
For fine-grained semantic correctness, we construct a QA pair $(q,a^\star)$ from the prompt and evaluate it with a frozen language model. 
Given token logits $\ell_t$ with $p_t=\softmax(\ell_t)$, the reward is the negated length-normalized cross-entropy plus margin objective:
\begin{equation}
\label{eq:r_llm}
\setlength{\abovedisplayskip}{6pt}
\setlength{\belowdisplayskip}{6pt}
\begin{aligned}
\scalebox{0.80}{$
R_{\text{vqa}}\!=\!-\!\tfrac{1}{T} \sum\limits_{t=1}^{T}\!\Big[
\log p_t[a_t^\star]
\!+\!\lambda_m\max\!\big(0,\,m\!-\!\ell_t[a_t^\star]\!+\!\max\limits_{u\!\ne\!a_t^\star}\ell_t[u]\big)
\Big].
$}
\end{aligned}
\end{equation}
To the best of our knowledge, this is the first work to integrate a differentiable VQA-based reward into inference-time controllable image generation and editing.

\subsection{Identity-Preserving KL Tether}
\label{subsec:kltether}
Strong reward guidance can cause the sampler to ``chase'' high reward values at the expense of the input’s identity, producing drift, layout distortion, or reward hacking.
To prevent this, we introduce an identity-preserving KL tether that softly pulls the predicted clean latent $\tilde \vz^{(k)}$
back toward the original latent representation $\vz_0$. Conceptually, this term corresponds to the gradient of a Kullback–Leibler divergence between 
the current clean-prediction distribution
$q(\tilde \vz|\vz^{(k)})$ and a reference Gaussian prior centered at the input latent $\vz_0$.
Minimizing this KL encourages the clean prediction to remain close to the input’s content and structure.
Taking the derivative of this KL with respect to latent  $\vz^{(k)}$ yields
\begin{equation}
\label{eq:kl}
\setlength{\abovedisplayskip}{6pt}
\setlength{\belowdisplayskip}{6pt}
g_{\text{KL},k} = -\lambda_{\text{KL}}\, J_{\FMDenoise}^\top \big(\tilde \vz^{(k)} - \vz_0\big)
\end{equation}
This regularizer moderates aggressive reward-driven updates, preventing drift, and anchoring the generation around the source identity and spatial layout.

\section{Experiments}
\label{sec:experiments}
We evaluate \modelnamenc using state-of-the-art diffusion backbones, Flux~\cite{blackforest2025flux}, Qwen~\cite{wu2025qwen}, and PixArt-$\alpha$~\cite{chen2024pixart} on two established benchmarks: \textsc{PIE-Bench} for image editing~\cite{ju2024pnp} and \textsc{T2I-CompBench} for compositional generation~\cite{huang2023t2i}. Additional details are provided in the Appendix.

\subsection{Image Editing Results}
\noindent \textbf{Quantitative Results.} We report quantitative editing performance in \Cref{tab:PIE_bench}, comparing \modelnamenc{} to a broad suite of open-source, training-free editing methods on PIE-Bench. All diffusion-based baselines use the same Flux backbone for a controlled comparison, with TurboEdit (SDXL) and AREdit (autoregressive) as exceptions. Under this shared backbone, Flux+\modelnamenc{} achieves consistent state-of-the-art performance. Relative to the strongest prior Flux-based baseline, Flux+\modelnamenc{} reduces Distance by $7.3\%$ (7.78 vs.\ 8.39) while keeping LPIPS within $6.0\%$ of the best Flux-based method (40.55 vs.\ 38.27), indicating better preservation of background and identity at comparable perceptual similarity. \modelnamenc also improves PSNR by $5.3\%$ (31.21 vs.\ 29.63) and SSIM by $2.6\%$ (89.67 vs.\ 87.44), yielding sharper and more structurally consistent reconstructions. For edit alignment, \modelnamenc{} increases Whole accuracy by $4.4\%$ (28.21$\rightarrow$29.44) and Edited accuracy by $8.6\%$ (24.51$\rightarrow$26.62), outperforming all Flux-based editors. These fidelity gains are obtained with competitive efficiency (43 NFEs and 20 sampling steps), corresponding to roughly $60$--$80\%$ fewer sampling steps than gradient-based editors that typically require 50--100 steps. In the few-step setting (4 sampling steps), \modelnamenc{} with Flux and Qwen Image further improves over prior fast editors (InstantEdit and TurboEdit), reducing Distance by up to $44.4\%$ and LPIPS by up to $25.8\%$, while increasing Whole and Edited accuracies by up to $11.1\%$ and $23.4\%$, respectively.

\begin{table}[t!]
\centering
\caption{\textbf{T2I compositional generation on \textsc{T2I-CompBench}.} 
Accuracy across fine-grained attribute binding (color, shape, texture), object relationships (spatial and non-spatial), and complex compositions. \modelnamenc consistently improves all base models (PixArt-$\alpha$, Flux, and Qwen Image). Best results are in \textbf{bold}.}
\vspace{-0.3cm}
\label{tab:t2icomp}
\resizebox{0.99\linewidth}{!}{
\begin{tabular}{lcccccc}
\toprule
\textbf{Model} & \textbf{Color} & \textbf{Shape} & \textbf{Texture} & \textbf{Spatial} & \textbf{Non-Spatial} & \textbf{Complex} \\
\midrule
SD v1.4~\cite{rombach2022high} & 0.38 & 0.36 & 0.42 & 0.12 & 0.31 & 0.31 \\
SD v2.1~\cite{rombach2022high} & 0.51 & 0.42 & 0.49 & 0.13 & 0.31 & 0.34 \\
SDXL~\cite{podell2023sdxl} & 0.64 & 0.54 & 0.56 & 0.20 & 0.31 & 0.41 \\
PixArt-$\alpha$~\cite{chen2024pixart} & 0.69 & 0.56 & 0.70 & 0.21 & 0.32 & 0.41 \\
DALL-E 2~\cite{openai2023dalle2} & 0.57 & 0.55 & 0.64 & 0.13 & 0.30 & 0.37 \\
DALL-E 3~\cite{openai2023dalle3} & 0.81 & 0.68 & 0.81 & {--} & {--} & {--} \\
\midrule
(1) PixArt-$\alpha$ DMD~\cite{yin2024one} & 0.38 & 0.34 & 0.47 & 0.19 & 0.30 & 0.36 \\
\rowcolor{gray!10} (1) + ReNO~\cite{eyring2024reno} & 0.64 & 0.57 & 0.72 & 0.25 & 0.31 & 0.46 \\
\rowcolor{rewardpurple!10} \textbf{(1) + \modelnamecolor} & \textbf{0.74} & \textbf{0.66} & \textbf{0.75} & \textbf{0.30} & \textbf{0.39} & \textbf{0.52} \\
\midrule
(2) Flux~\cite{blackforest2025flux} & 0.75 & 0.61 & 0.69 & 0.26 & 0.33 & 0.47 \\
\rowcolor{gray!10} (2) + ReNO~\cite{eyring2024reno} & 0.81 & 0.64 & 0.72 & 0.29 & 0.35 & 0.49 \\
\rowcolor{rewardpurple!10} \textbf{(2) + \modelnamecolor} & \textbf{0.88} & \textbf{0.69} & \textbf{0.78} & \textbf{0.33} & \textbf{0.42} & \textbf{0.57} \\
\midrule
(3) Qwen Image~\cite{wu2025qwen} & 0.83 & 0.72 & 0.80 & 0.35 & 0.39 & 0.61 \\
\rowcolor{gray!10} (3) + ReNO~\cite{eyring2024reno} & 0.84 & 0.75 & 0.84 & 0.36 & 0.43 & 0.63 \\
\rowcolor{rewardpurple!10} \textbf{(3) + \modelnamecolor}& \textbf{0.91} & \textbf{0.83} & \textbf{0.90} & \textbf{0.39} & \textbf{0.51} & \textbf{0.78} \\
\bottomrule
\end{tabular}
}
\end{table}

\begin{table}[t]
\centering
\caption{\textbf{Ablation on key \modelnamenc components.}}
\vspace{-0.3cm}
\label{tab:ablcompo}
\renewcommand{\arraystretch}{0.90}
\resizebox{0.99\linewidth}{!}{
\begin{tabular}{lccccccc}
\toprule
\textbf{Setting / Variant} &
\makecell{\textbf{Distance} $\downarrow$\\($\times10^3$)} &
\textbf{PSNR} $\uparrow$ &
\makecell{\textbf{LPIPS} $\downarrow$\\($\times10^3$)} &
\makecell{\textbf{MSE} $\downarrow$\\($\times10^4$)} &
\makecell{\textbf{SSIM} $\uparrow$\\($\times10^2$)} &
\textbf{Whole} $\uparrow$ &
\textbf{Edited} $\uparrow$ \\
\midrule

\xmark~ Dynamic Reward Weighting. & 8.47 & 30.77 & 38.78 & 34.67 & 89.37 & 29.02 & 27.01 \\
\xmark~ Semantic Primitives (SPs) & 9.03 & 31.19 & 39.12 & 35.01 & 88.47 & 27.45 & 26.51 \\
\xmark~  Reward-Aware Step Size & 9.15 & 31.38 & 39.56 & 35.21 & 89.12 & 28.32 & 26.92 \\
\xmark~  KL-Tether & 9.56 & 29.98 & 40.26 & 35.28 & 87.23 & 27.98 & 26.13 \\
\rowcolor{rewardpurple!10} \textbf{\modelnamecolor{} (Full)} & \textbf{7.64} & \textbf{32.09} & \textbf{38.47} & \textbf{33.57} & \textbf{90.21} & \textbf{29.78} & \textbf{27.57} \\
\bottomrule
\end{tabular}
}
\vspace{-0.3cm}
\end{table}

\noindent \textbf{Qualitative Results.}
\Cref{fig:main_fig} shows that \modelnamenc produces edits that are both more instruction-faithful and spatially precise than baselines. In the viewpoint transformation of the sculpture, \modelnamenc{} changes the perspective without introducing major distortions compared to baselines. 
For object replacement tasks, such as replacing the surfboards with flowers or changing the fruits to pizza, competing methods often under-edit, over-edit, or generate implausible replacements, whereas \modelnamenc{} performs the intended substitution cleanly while maintaining the surrounding layout and appearance. Similarly, in the animal replacement example, baselines such as InfEdit and InstantEdit fail to replace the tiger or cannot adhere to the specified attributes,  while \modelnamenc{} generates the brown cat with coherent shape, color, and pose. In relational or compositional edits, such as making the parrots kiss or changing the forest into a wizard house, \modelnamenc{} again yields outputs that better reflect the target semantics without the leakage, structural drift, or background corruption observed in other methods.

\subsection{Image Generation Results}
\noindent \textbf{Quantitative Results.} 
\Cref{tab:t2icomp} reports text-to-image generation performance on \textsc{T2I-CompBench} across six categories spanning fine-grained attribute binding, object relationships, and complex multi-constraint prompts. Across all three base models (PixArt-$\alpha$, Flux, and Qwen Image), \modelnamenc consistently improves composition accuracy and outperforms the training-free reward-based baseline ReNO across every category. For Flux, \modelnamenc improves overall performance by approximately 12.5\%, while for Qwen Image, the improvement reaches 12.8\%, with particularly strong boosts in non-spatial and complex compositional categories. These results underscore \modelnamenc{}’s effectiveness in enhancing attribute binding and relational coherence across diverse generation backbones.

\begin{figure}[t!]
\includegraphics[width=0.99\linewidth]{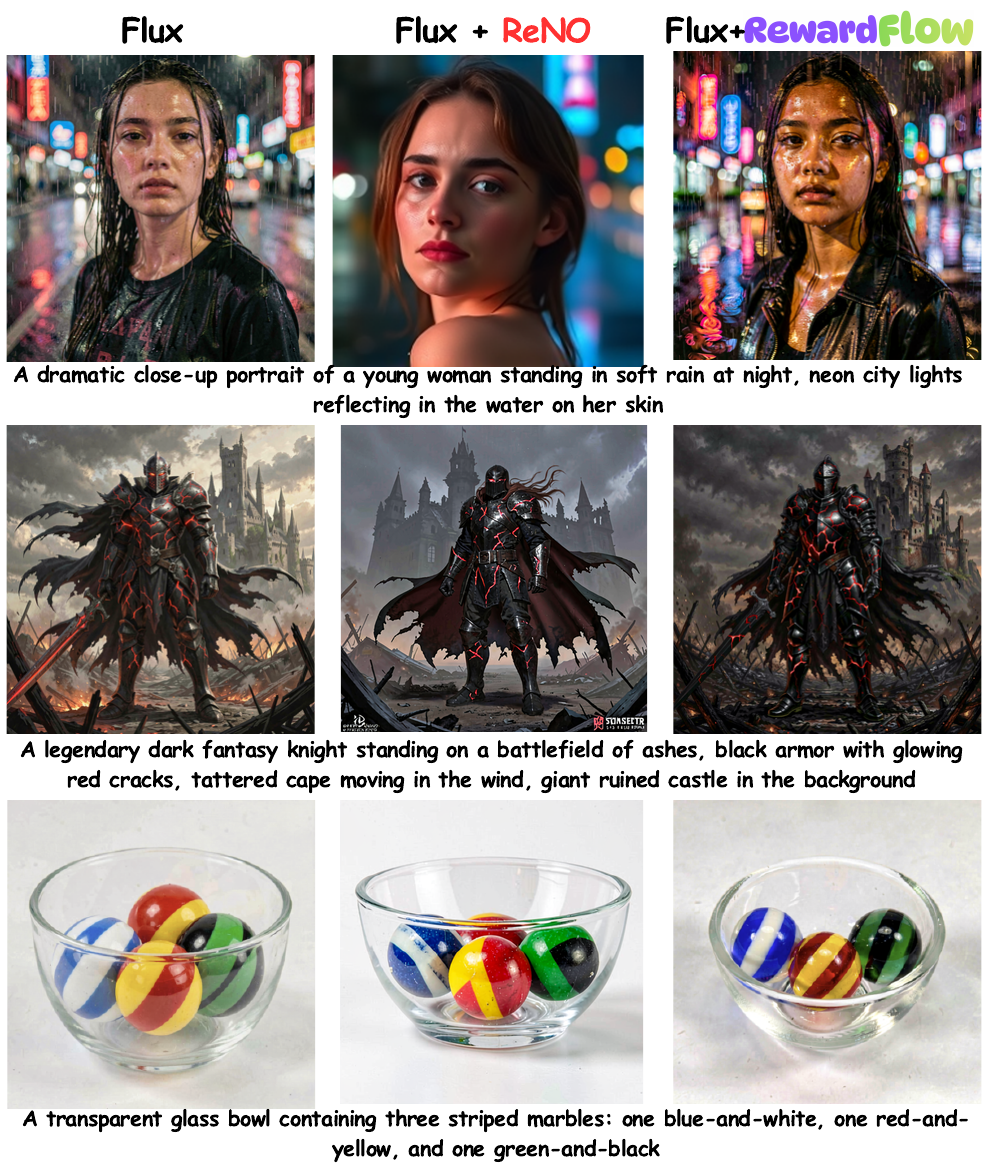}
\vspace{-0.4cm}
\caption{\textbf{Text-to-image qualitative results.} Across all prompts, \modelnamenc produces images that exhibit higher alignment with the textual descriptions while also generating outputs with more visually appealing composition and aesthetics. }
\label{fig:t2iqual}
\vspace{-0.2cm}
\end{figure}
\begin{figure}[t!]
\centering
\includegraphics[width=0.99\linewidth]{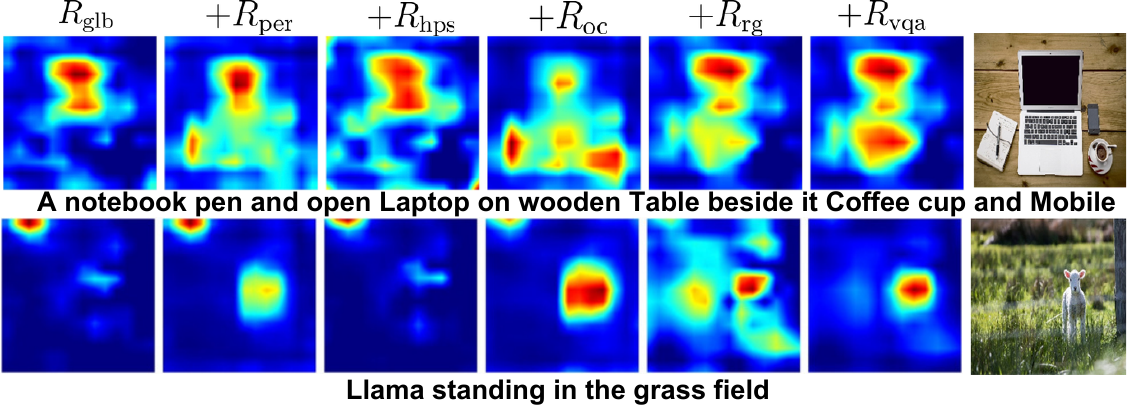}
\vspace{-0.4cm}
\caption{\textbf{Gradient localization across reward combinations.} Including all rewards concentrates gradients to accurate object contours and eliminates leakage.
}
\label{fig:gradmaps}
\vspace{-0.3cm}
\end{figure}

\noindent \textbf{Qualitative Results.} 
\Cref{fig:t2iqual} compares examples generated from ReNO~\cite{eyring2024reno} and \modelnamenc{} with Qwen Image backbone. 
Across diverse prompts, \modelnamenc produces images with stronger semantic alignment and improved aesthetic quality. Compared to the base model and ReNO, our method consistently enhances color vibrancy, local detail, and prompt adherence. In the portrait example, \modelnamenc{} better captures the neon reflections and wet-skin appearance, while in the fantasy knight example, \modelnamenc{} generates a more coherent dark atmosphere, sharper armor details, and the ruined castle that other methods fail to realize. In the striped marbles example,  \modelnamenc{} more faithfully satisfies the specified object count and patterns. 

\begin{figure}[t!]
\includegraphics[width=0.99\linewidth]{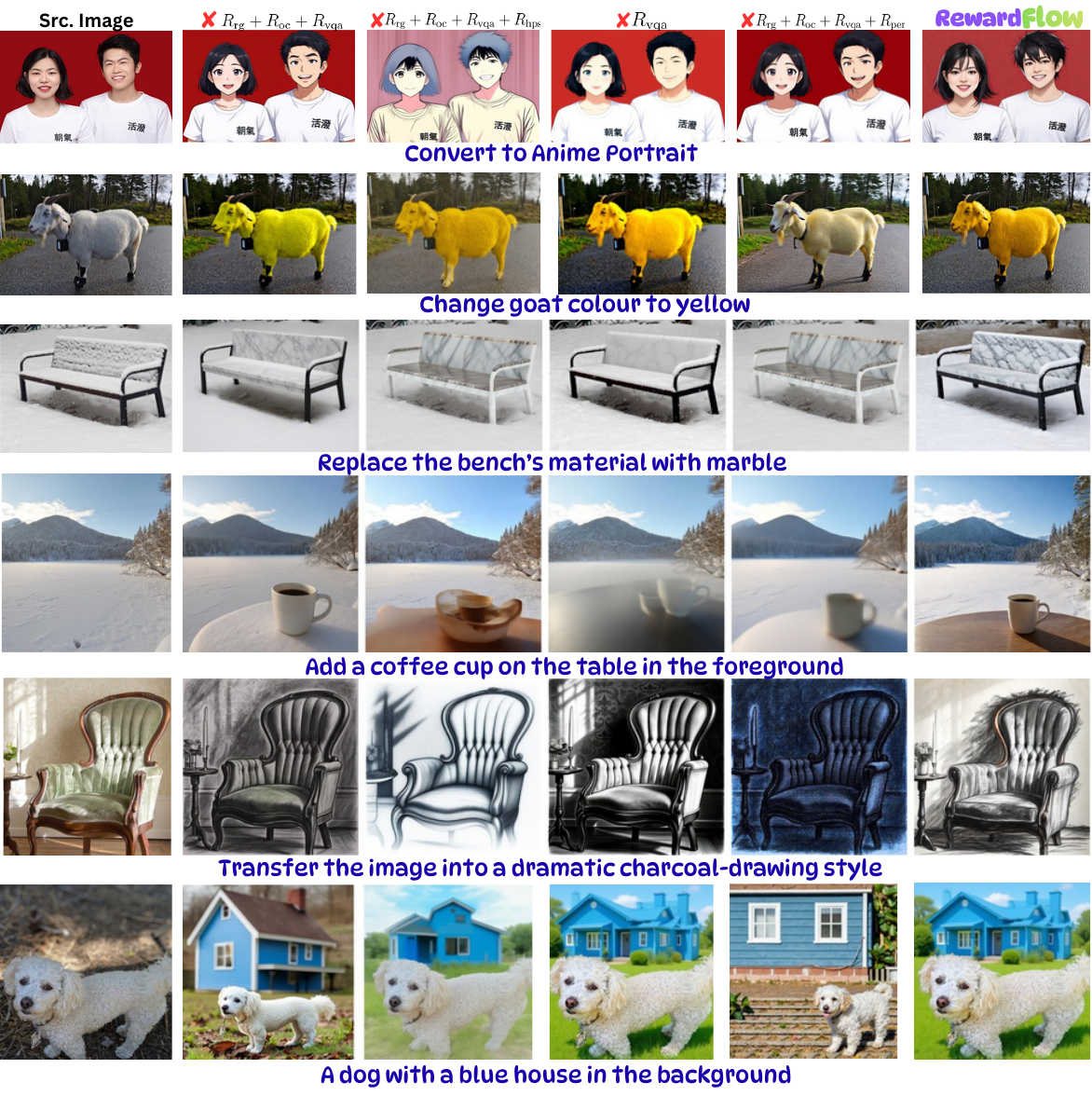}
\vspace{-0.4cm}
    \caption{\textbf{Effect of removing reward components.} 
    (\xmark~RC, SAM, LLM, HPS, and PE) denote excluding ($R_{\text{rg}}$, $R_{\text{oc}}$, $R_{\text{vqa}}$, $R_{\text{hps}}$, and $R_{\text{per}}$), respectively. 
\modelnamenc{} (all rewards) achieves semantically precise edits, modifying only instruction-relevant content while maintaining background and context integrity.}
    \label{fig:reward-ablation}
    \vspace{-0.4cm}
\end{figure}

\section{Ablation Studies}

\noindent \textbf{Reward Components.} We analyze both quantitative and qualitative effects of each reward component and visualize their gradient localization behavior.
As shown in \Cref{fig:gradmaps} and \Cref{tab:ablationrewards}, with only the global alignment reward ($R_{\text{glb}}$, first column), gradients are broadly distributed, yielding weak spatial focus (Distance 11.23, SSIM 84.09). Including the perceptual reward ($R_{\text{per}}$) sharpens local structure (LPIPS 45.22 → 43.12), slightly improving focus around relevant regions. The human preference reward ($R_{\text{hps}}$) improves realism and global coherence (PSNR 27.57 → 28.82), reducing perceptual noise and producing smoother tones in tasks such as ``marble bench'' and portrait stylization (\Cref{fig:reward-ablation}). The object consistency reward ($R_{\text{oc}}$) improves gradient concentration around intended objects (\Cref{fig:gradmaps}), reducing leakage and improving spatial precision (Distance 9.77 → 8.39). This effect is clearly seen in localized edits such as ``yellow goat'' and ``blue house dog'' (\Cref{fig:reward-ablation}).
Adding the region-level grounding reward ($R_{\text{rg}}$) further localizes gradients to prompt-relevant regions (PSNR 29.44 → 30.12, Whole 21.47 → 26.47), ensuring edits such as the ``coffee cup'' placement occur in the correct area without disturbing context (\Cref{fig:reward-ablation}). Finally, the VQA reward ($R_{\text{vqa}}$) provides the strongest fine-grained supervision (PSNR 32.09, SSIM 90.21), producing sharply focused gradient activations that align with object contours and yield semantically precise results across all tasks.
Overall, global rewards ($R_{\text{glb}}, R_{\text{per}}, R_{\text{hps}}$) ensure semantic and perceptual coherence, while localized rewards ($R_{\text{oc}}, R_{\text{rg}}, R_{\text{vqa}}$) progressively concentrate gradients to target regions, achieving precise instruction-faithful edits. 

\begin{figure}[t!]
\includegraphics[width=0.99\linewidth]{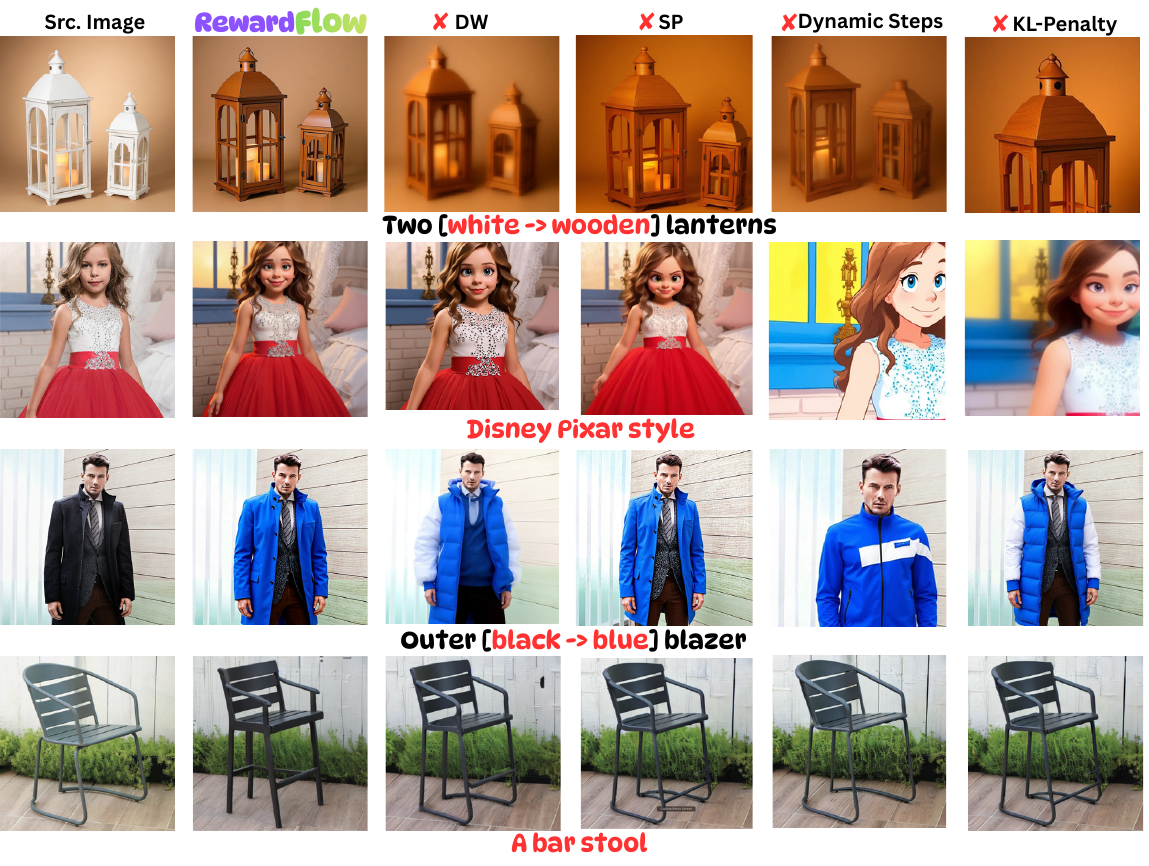}
\vspace{-0.4cm}
    \caption{\textbf{Ablations illustrating the effect of removing key components.}
\modelnamenc (all components) achieves the best visual consistency and instruction alignment.}
    \label{fig:method-ablation}
    \vspace{-0.2cm}
\end{figure}

\noindent \textbf{RewardFlow Method Components.} In Table~\ref{tab:ablcompo} and Figure~\ref{fig:method-ablation}, we ablate each component of our adaptive policy to assess its contribution. The full model achieves the best results (Distance 7.64, PSNR 32.09, SSIM 90.21), confirming the effectiveness of jointly adaptive control. Removing dynamic reward weighting reduces PSNR by 1.32 and SSIM by 0.84, as fixed weights fail to adapt to evolving reward satisfaction—evident in the “two [white$\rightarrow$wooden] lanterns” edit, where color shifts occur but texture consistency degrades. Without semantic primitives (SPs), Distance rises to 9.03 and Whole drops by 2.33, leading to interference between objectives and inconsistent stylization, as seen in the “disney pixar style” example. Fixing the step size worsens Distance (9.15) and PSNR (31.38), producing unstable updates such as uneven recoloring in the “outer [black$\rightarrow$blue] blazer” edit. Excluding the KL tether causes the most severe degradation (PSNR -2.11, SSIM -1.89) and structural drift, exemplified by distortions in the bar stool geometry. Overall, SPs enable disentangled control, dynamic weighting maintains balanced optimization, adaptive steps ensure stable convergence, and the KL tether preserves structural fidelity, collectively supporting coherent and precise controllable generation.

\begin{table}[t]
\centering
\caption{\textbf{Ablation on reward components.} Each column indicates whether the corresponding reward is enabled (\cmark) or disabled (\xmark). 
}
\vspace{-0.3cm}
\label{tab:ablationrewards}
\setlength{\tabcolsep}{3pt}
\resizebox{0.99\linewidth}{!}{
\begin{tabular}{ccccccccccccc}
\toprule
{$R_{\text{glb}}$} & 
{$R_{\text{per}}$} & 
{$R_{\text{hps}}$} & 
{$R_{\text{oc}}$} & 
$R_{\text{rg}}$ &
{$R_{\text{vqa}}$} &
\makecell{\textbf{Distance} $\downarrow$ \\ ($\times10^3$)} &
\textbf{PSNR} $\uparrow$ &
\makecell{\textbf{LPIPS} $\downarrow$ \\ ($\times10^3$)} &
\makecell{\textbf{MSE} $\downarrow$ \\ ($\times10^4$)} &
\makecell{\textbf{SSIM} $\uparrow$ \\ ($\times10^2$)} &
\textbf{Whole} $\uparrow$ &
\textbf{Edited} $\uparrow$ \\
\midrule
 \cmark & \xmark & \xmark & \xmark & \xmark & \xmark & 11.23 & 26.33 & 45.22 & 39.34 & 84.09 & 19.33 & 21.22 \\
\cmark & \cmark & \xmark & \xmark & \xmark & \xmark & 10.21 & 27.57 & 43.12 & 37.45 & 85.17 & 19.88 & 22.85 \\
\cmark & \cmark & \cmark & \xmark & \xmark & \xmark &  9.77 & 28.82 & 41.67 & 37.12 & 86.39 & 20.12 & 23.33 \\
 \cmark & \cmark & \cmark & \cmark & \xmark & \xmark &  8.39 & 29.44 & 40.58 & 36.55 & 87.71 & 21.47 & 24.75 \\
\cmark & \cmark & \cmark & \cmark & \cmark & \xmark &  8.01 & 30.12 & 40.02 & 35.28 & 88.92 & 26.47 & 25.91 \\
\rowcolor{rewardpurple!30} \cmark & \cmark & \cmark & \cmark & \cmark & \cmark &  \textbf{7.64} & \textbf{32.09} & \textbf{38.47} & \textbf{33.57} & \textbf{90.21} & \textbf{29.78} & \textbf{27.57} \\
\bottomrule
\end{tabular}
}
\vspace{-0.3cm}
\end{table}

\section{Conclusion}\label{sec:conclusion}
We introduce \modelnamecolor{}, a training-free framework that steers pretrained text-guided image editing and generation models using multi-reward Langevin dynamics. By combining global, localized, and VQA-based rewards with a prompt-aware adaptive policy and a KL tether, \modelnamenc{} achieves fine-grained, spatially precise control while preserving identity and layout. 
Extensive experiments demonstrate consistent improvements in edit fidelity, compositional alignment, and generation quality over strong training-free baselines. We believe treating controllable generation as reward-guided sampling offers a general test-time alignment strategy, with promising extensions to video editing.

\section*{Acknowledgments}
This research was partially supported by Google, the Google TPU Research Cloud (TRC) program, the U.S. Defense Advanced Research Projects Agency (DARPA) under award HR001125C0303, and the U.S. Army under contract W5170125CA160. The views and conclusions contained herein are those of the authors and should not be interpreted as necessarily representing the official policies, either expressed or implied, of Google, DARPA, the U.S. Army, or the U.S. Government. The U.S. Government is authorized to reproduce and distribute reprints for governmental purposes notwithstanding any copyright annotation therein.\\
{
    \small
    \bibliographystyle{ieeenat_fullname}
    \bibliography{main}
}

\clearpage
\setcounter{page}{1}

\twocolumn[{%
\renewcommand\twocolumn[1][]{#1}%
\maketitlesupplementary
\centering
 \includegraphics[height=.8\textheight,keepaspectratio]{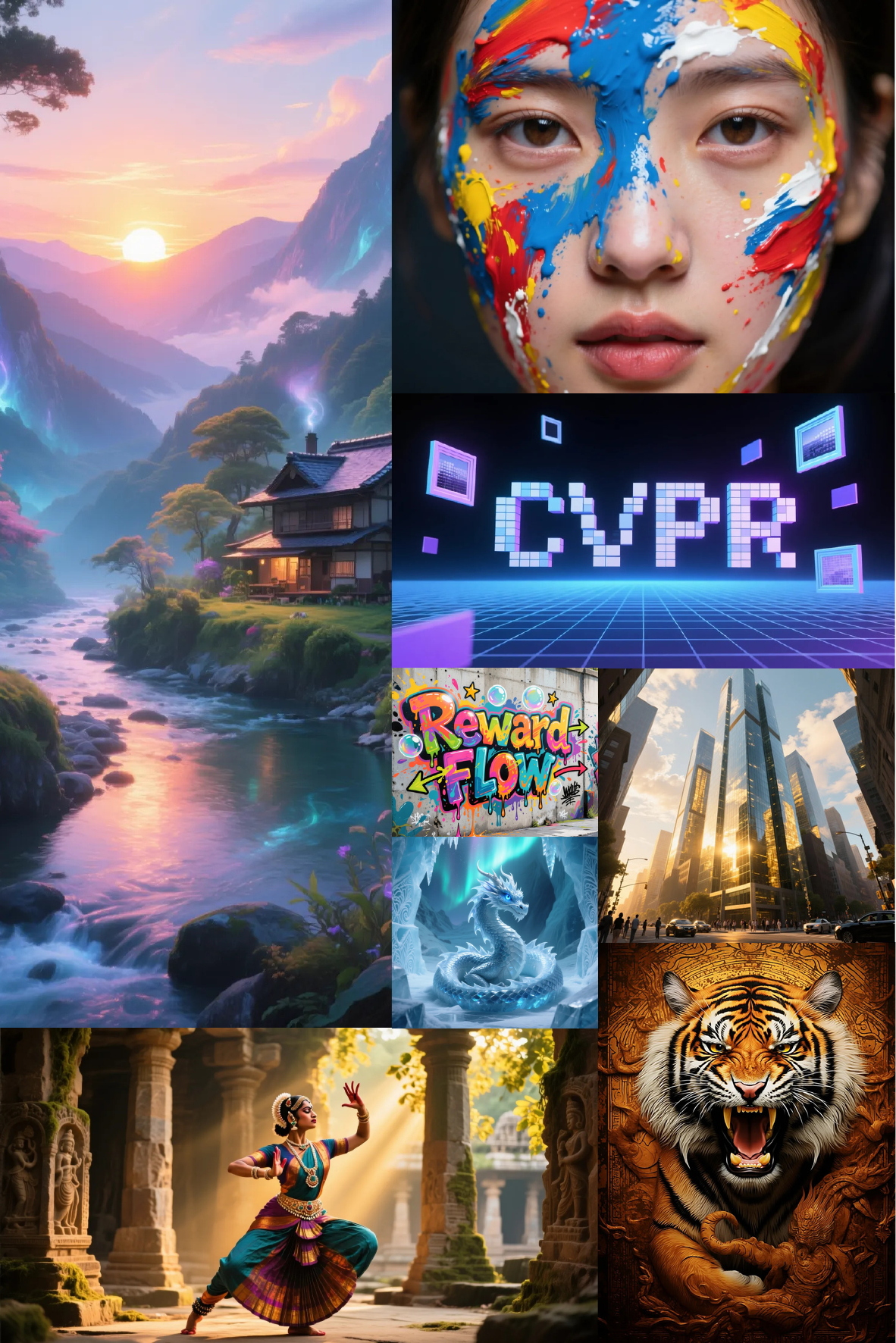}
\captionof{figure}{\textbf{High-resolution images generated by \modelnamecolor{}.}}
\label{fig:fullpage_image}
\vspace{0.2cm}
}]

\clearpage
\newpage
\section{SDE Formulation}
\label{sec:sde-proof}
In this section, we detail the stochastic differential equation (SDE) that grounds the Langevin-style reverse update in \cref{eq:update}, specify the diffusion-strength schedule~$\gamma_k$, and provide a derivation showing how \cref{eq:update} emerges from sampling a prompt-tilted latent density.

\noindent \textbf{Prompt-tilted target density.}
Let $q_t(\vz \mid \vp)$ denote the unconditional latent distribution at time $t$ for prompt $\vp$.
Given total reward $R_{\mathrm{tot}}(\vz,t,\vp)$ obtained by combining the differentiable rewards and
KL potential
\begin{equation}
\setlength{\abovedisplayskip}{6pt}
\setlength{\belowdisplayskip}{6pt}
K(\vz,t;\vz_0)
:= \tfrac{1}{2}\,\bigl\|\tilde{\vz}(\vz,t,\vp) - \vz_0\bigr\|_2^2,
\label{prompttileeq}
\end{equation}
we define the \emph{prompt-tilted} target density
\begin{equation}\label{eq7}
\setlength{\abovedisplayskip}{6pt}
\setlength{\belowdisplayskip}{6pt}
\scalebox{0.8}{$
\rho_t(\vz \mid \vp, x)
\propto
q_t(\vz \mid \vp)\,
\exp\!\Big(
    \lambda_R\, R_{\mathrm{tot}}(\vz,t,\vp)
    \!-\!
    \lambda_{\mathrm{KL}}\, K(\vz,t;\vz_0)
\Big).
$}
\end{equation}
where $\vz_0 = \mathrm{Enc}(\vx)$ denotes the clean latent of the (optional)
source image $\vx$. Taking the gradient of the log-density in \cref{eq7} yields 
\begin{equation*}
\setlength{\abovedisplayskip}{6pt}
\setlength{\belowdisplayskip}{6pt}
    \begin{aligned}
    \nabla_\vz \log \rho_t(\vz \mid p,x)
    &=\nabla_\vz \log q_t(\vz \mid p)
    \\
    &\quad + \lambda_R\,\nabla_\vz R_{\mathrm{tot}}(\vz,t,\vp)
    \\
    &\quad - \lambda_{\mathrm{KL}}\,\nabla_\vz K(\vz,t;\vz_0).
    \end{aligned}
\end{equation*}

Each reward $R_i(I,p)$ is defined in image space. For step $k$, let
$g_{\vI,i}^{(k)} := \nabla_{\vI} R_i(I^{(k)},p)$ denote the image-space
gradient. Using the decoder and denoiser Jacobians, the reward drift in
 \cref{eq:prelim-reward-drift} can be written as
\begin{equation}
\setlength{\abovedisplayskip}{6pt}
\setlength{\belowdisplayskip}{6pt}
g_{R_i,k}
=
\lambda_R\,
J_{\mathrm{Den}}(\vz^{(k)},t_k,\vp)^{\!\top}
J_{\mathrm{Dec}}(\tilde \vz^{(k)})^{\!\top}\,
g_{\vI,i}^{(k)} .
\end{equation}
Summing over rewards then yields the fused reward drift
$g_{R\mathrm{tot},k} = \lambda_R \nabla_{\vz^{(k)}} R_{\mathrm{tot}}(\vz^{(k)},t_k,p)$. 
In addition, differentiating $K(\vz^{(k)},t_k;\vz_0)$ with respect to $\vz^{(k)}$ yields \cref{eq:kl}, so
$g_{\mathrm{KL},k} = -\lambda_{\mathrm{KL}}\,\nabla_{\vz^{(k)}}
K(\vz^{(k)},t_k;\vz_0)$.

\subsection{Langevin SDE and Discrete Update}
We introduce an \emph{algorithmic time} variable $s \in [0,S]$ and a
monotone schedule $t(s)$ from algorithmic time to diffusion time, with
$t(0) = \bar{t}$ and $t(S) = 0$. We consider the overdamped Langevin
SDE whose stationary distribution at each $t$ is the prompt-tilted
density $\rho_t$:
\begin{equation}\label{eq:s2}
\setlength{\abovedisplayskip}{6pt}
\setlength{\belowdisplayskip}{6pt}
    \mathrm{d}\vz_s
    \;=\;
    \nabla_\vz \log \rho_{t(s)}\!\bigl(\vz_s \mid p,x\bigr)\,\mathrm{d}s
    \;+\;
    \sqrt{2\,\gamma(s)}\,\mathrm{d}W_s,
\end{equation}
where $W_s$ is standard Brownian motion and $\gamma(s)>0$ controls the
diffusion strength.

Let $s_0 < s_1 < \dots < s_K$ be a discretization of $[0,S]$ with step
sizes $\eta_k\!=\!s_{k+1} - s_k$, and write $t_k\!=\!t(s_k)$,
$\gamma_k\!=\!\gamma(s_k)$, and $\vz^{(k)} \approx \vz_{s_k}$. Applying
Euler--Maruyama to \cref{eq:s2} yields
\begin{equation}
\setlength{\abovedisplayskip}{6pt}
\setlength{\belowdisplayskip}{6pt}
    \begin{aligned} 
    \vz^{(k+1)}
    &= \vz^{(k)}
    + \eta_k\,
      \nabla_\vz \log \rho_{t_k}\!\bigl(\vz^{(k)} \mid p,x\bigr)
      \\
      &\quad
    + \sqrt{2\,\gamma_k \eta_k}\,\xi_k,
    \qquad
    \xi_k \sim \mathcal{N}(0,\vI).
    \end{aligned}
    \label{s3}
\end{equation}
Substituting Eq.~\eqref{eq7} and using flow-matching score approximation $v_\theta(\vz,t,\vp) \approx\
\nabla_\vz \log q_t(\vz \mid p)$ we obtain
\begin{equation}
\setlength{\abovedisplayskip}{6pt}
\setlength{\belowdisplayskip}{6pt}
    \begin{aligned}
    \nabla_\vz \log \rho_t(\vz \mid p,x)
    &\approx\
    v_\theta(\vz,t,\vp)
    + g_{R_\mathrm{tot}}(\vz,t,\vp)
    \\
    &\quad
    + g_{\mathrm{KL}}(\vz,t;\vz_0).
    \end{aligned}
\end{equation}
Evaluating this at $(\vz^{(k)},t_k)$ in \cref{s3} yields
\begin{equation}
\setlength{\abovedisplayskip}{6pt}
\setlength{\belowdisplayskip}{6pt}
    \begin{aligned}
    \vz^{(k+1)}
    &= \vz^{(k)}
    + \eta_k
      \Bigl(
        f_k + g_{R\mathrm{tot},k} + g_{\mathrm{KL},k}
      \Bigr) + \xi_k,
    \\
    &\quad
    \qquad
    \xi_k \sim \mathcal{N}\bigl(0, 2\gamma_k \eta_k \vI\bigr),
    \end{aligned}
\end{equation}
where $f_k := v_\theta(\vz^{(k)},t_k,\vp)$ is the backbone drift. This is
exactly the stochastic update stated in \cref{eq:update} of the main paper, now
seen as an Euler--Maruyama discretization of the Langevin SDE in
\cref{s3} targeting the prompt-tilted density in
\cref{eq7}, with the reward terms defining the controllability potential and the KL tether stabilizing identity and layout.

\subsection{Noise variance schedule \texorpdfstring{$\gamma_k$}{gamma\_k}}

The Gaussian perturbation in \cref{eq:update} is parameterized by the
time-dependent variance $\gamma_k$. We instantiate $\gamma_k$ with a
monotonically decreasing schedule
\begin{equation}
\setlength{\abovedisplayskip}{6pt}
\setlength{\belowdisplayskip}{6pt}
    \begin{aligned}
    \gamma_k
    &= \gamma_{\min}
    + \bigl(\gamma_{\max} - \gamma_{\min}\bigr)
      \left(\frac{t_k}{\bar{t}}\right)^{\rho},
    \\
    &\text{where }\gamma_{\min},\gamma_{\max}>0,\ \rho>0,
    \end{aligned}
    \label{d4}
\end{equation}
so that early steps at high noise levels ($t_k \approx \bar{t}$) use larger diffusion (exploration), while late steps near $t_k \approx 0$ use smaller diffusion, focusing on refinement. In the special case
$\gamma_{\min}\!=\!\gamma_{\max}$, \cref{d4} reduces to a constant-noise Langevin sampler.\looseness-1

\section{Datasets and Evaluation}

To ensure a fair comparison, we adopt the same evaluation protocols and metrics as defined in the original papers of each respective dataset.

\noindent \textbf{\textsc{T2I-CompBench.}}
\textsc{T2I-CompBench} is a large-scale benchmark designed to evaluate compositional text-to-image generation in open-world settings, and consists of approximately 6,000 prompts categorized into three key tasks: \textit{attribute binding}, \textit{object relationships}, and \textit{complex compositions}. Each prompt describes scenes with multiple objects and attributes, requiring precise alignment between textual semantics and visual structure. For evaluation, the benchmark employs a set of compositional metrics. Attribute binding is assessed using BLIP-based VQA that queries each object’s attribute independently (\eg``What color is the bench?''). Spatial relations are evaluated using UniDet, a pre-trained object detector, to check the relative positioning of objects via bounding box analysis. For complex scenes, a compositional consistency score is computed by aggregating CLIPScore, BLIP-VQA accuracy, and UniDet spatial relation correctness. This framework enables a detailed understanding of how well models handle fine-grained compositional constraints beyond conventional image-text similarity.\looseness-1

\noindent \textbf{\textsc{GenEval.}}
\textsc{GenEval} is a structured evaluation suite targeting fine-grained text-to-image alignment at the object level. It introduces prompts designed to probe a model's ability to generate images with correct object \textit{presence}, \textit{co-occurrence}, \textit{counting}, \textit{spatial arrangement}, and \textit{color attribution}. Each generated image is evaluated using automated pipelines based on pre-trained vision models. Object detectors verify the existence and number of instances for specified entities, while spatial metrics assess whether objects appear in the correct geometric configuration (\eg left/right or above/below). Color attributes are checked by segmenting object regions and comparing predicted colors with prompt specifications. Each task yields binary correctness judgments, and the results are reported as per-category accuracies along with an overall compositional accuracy score. GenEval has been shown to correlate strongly with human judgments and helps isolate specific failure modes such as incorrect object counts or attribute swaps.

\noindent \textbf{\textsc{PIE-Bench.}}
\textsc{PIE-Bench} is a comprehensive benchmark for evaluating text-guided image editing systems. It comprises 700 real-world and artistic images, each paired with a \textit{source prompt}, a \textit{target prompt}, a natural language \textit{editing instruction}, and a binary \textit{editing mask}. The edits are drawn from ten categories, including object addition, removal, replacement, attribute changes (\eg color, pose), material substitution, background edits, and global style transformations. The benchmark evaluates two core criteria: (1) \textbf{Edit Fidelity}, which measures how well the edited image aligns with the target prompt, typically using CLIPScore or similar semantic similarity metrics; and (2) \textbf{Content Preservation}, which assesses how much of the non-edited image content remains unchanged, computed via PSNR or SSIM on unmasked (non-edit) regions. PIE-Bench allows for quantitative and targeted assessment of how effectively models perform localized or global edits while preserving image realism and structure.

\begin{figure}[t!]
    \centering
    \includegraphics[width=0.99\linewidth]{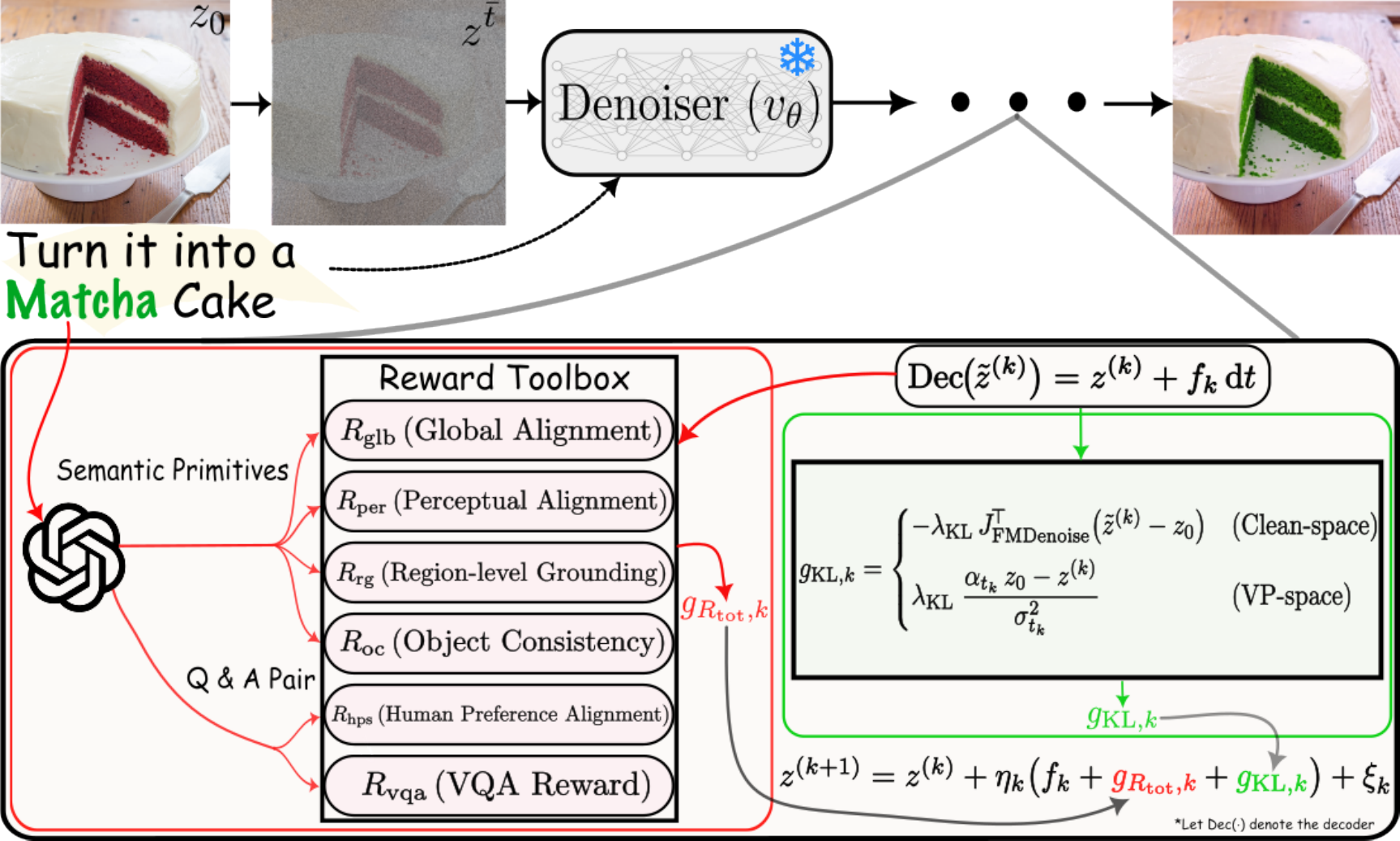}
    \vspace{-0.3cm}
    \caption{
    \textbf{Overview of the \modelnamenc{} framework.}
    }
    \label{fig:overview}
    \vspace{-0.3cm}
\end{figure}

\section{Implementation Details}
\label{sec:implementation}
In this section, we provide additional implementation details for \modelnamenc{}. An overview of the method is illustrated in \Cref{fig:overview}. Unless otherwise stated, we use the same hyperparameters across all backbones, datasets, and tasks.
All experiments are run on a single node with $2\times$ NVIDIA A100 GPUs (80\,GB each). We implement \modelnamenc{} in PyTorch with automatic mixed precision (AMP) for all backbones and reward networks, which reduces memory footprint and latency without affecting visual quality. Unless otherwise noted, we use a batch size of $1$ per GPU for editing experiments and $2$ for text-to-image generation.

\noindent \textbf{Backbones and Resolution.}
We instantiate \modelnamenc{} on three pretrained flow-matching / diffusion backbones: PixArt-$\alpha$, Flux,
and a Qwen-based latent diffusion model. All images are generated and edited at $1024\times1024$ resolution. We use the official checkpoints and sampling schedules for each backbone and do not fine-tune any model weights; \modelnamenc{} operates purely at inference time.
For image editing, given a prompt $\vp$ and source image $\vx$, we encode the image into a clean latent $\vz_0\!=\!\Enc(\vx)$, initialize a noisy latent $\vz^{(0)}$ at a fixed noise level $\bar{t}$ as in the backbone, and run $K\!=\!35$ reverse steps following the update in \cref{eq:update}. For unconditional text-to-image generation, $\vz_0$ is sampled from the backbone’s prior and the KL tether is disabled ($\lambda_{\mathrm{KL}}\!=\!0$).\looseness-1

\noindent \textbf{Prompt Parsing and Semantic Primitives.}
Before sampling, we parse each prompt $\vp$ once using GPT-5 to extract:
\begin{itemize}
  \item A set of {Semantic Primitives} 
  $\mathrm{SP}(\vp)\!=\!\{p_m\}_{m=1}^M$, where each $p_m$ is a
  short, atomic instruction (\eg ``remove cap'', ``add sunglasses'', \etc).
  \item A small set of VQA pairs $\{(q_j,a_j^\star)\}_{j=1}^{J_{\text{vqa}}}$ that probe
  fine-grained aspects of the intended edit (\eg
  ``What is on the person's head?'' $\rightarrow$ ``Nothing'').
\end{itemize}
As shown in the prompt template in \Cref{fig:vqaprompt}, we instruct the model to ensure that each SP is self-contained and that the VQA questions are answerable from the final image without ambiguity. This one-time parsing step is performed offline and cached for all subsequent sampling runs with the same prompt. For multi-instruction prompts, SPs prevent interference between unrelated objectives and enable per-primitive reward computation.

\begin{figure*}[t!]
\centering
\resizebox{0.99\textwidth}{!}{
\begin{planbox}{Vision-language Editing Assistant.} 
\small
You are a vision-language assistant. 
You receive an image and a short edit instruction.

\vspace{0.1cm}

\textbf{1) Extract short edit prompts:} output a compact list of 5--12 atomic, actionable tags/phrases that guide the image edit. 

Include:
 \begin{itemize}[itemsep=0em, leftmargin=2.5em]
   \item Visible subject descriptors (pose, angle, clothing items) actually present.
   \item The edit action(s) and key visual attributes (style, color, size, placement).
   \item Constraints to preserve identity, lighting, composition, realism, and continuity.
   \item Any practical rendering notes (alignment, shadows, reflections, edges).
\end{itemize}

\textbf{2) Create exactly one Q\&A pair focused on the final edited image's appearance.}

 \begin{itemize}[itemsep=0em, leftmargin=2.5em]

    \item Ask **one** question that would most affect the final look (\eg style, colorway, size/scale, placement, material/finish, mood/lighting continuity).
   
    \item Give **one** concise answer based on the image/instruction; if not determinable, answer "Unspecified from image."
\end{itemize}

\vspace{0.1cm}
\textbf{\#\# Rules}
\begin{itemize}[itemsep=0em, leftmargin=2.5em]
\item[-] **Output JSON only** in the exact schema below---no extra text.
\item[-] Keep each short prompt <= 6 words; imperative, neutral wording.
\item[-] Do not invent details not visible or implied by the instruction.
\item[-] Avoid sensitive inferences (\eg ethnicity, health, etc.).
\item[-] American English.
\end{itemize}

\vspace{0.1cm}
\textbf{\#\# Input}

EDIT\_INSTRUCTION:  \texttt{{\{edit\_instruction\}}}

\vspace{0.1cm}
\textbf{\#\# Output schema (JSON only)}

{\ttfamily
\hspace*{0em}\{\\
\hspace*{0.5em}"short\_prompts": ["<tag1>", "<tag2>", "..."],\\
\hspace*{0.5em}"qna": \{\\
\hspace*{2em}"question": "<visual-outcome question>",\\
\hspace*{2em}"answer": "<concise answer or 'Unspecified from image'>"\,\\
\hspace*{2em}\}\\
\hspace*{0em}\}
}
\end{planbox}
}
\vspace{-0.3cm}
\caption{\textbf{Prompt template} used for semantic primitives and $\mathcal{R}_{\text{vqa}}$.}
\label{fig:vqaprompt}
\end{figure*}

\noindent \textbf{Rewards and Feature Extractors.}
At every denoising step, each reward is evaluated on $\vI^{(k)}$ and the corresponding SPs, producing both a scalar score and an image-space gradient. We briefly summarize implementation choices for each.

\noindent \textbf{Global and perceptual rewards ($R_{\mathrm{glb}}$, $R_{\mathrm{per}}$).}
For the global semantic reward $R_{\mathrm{glb}}$ we use a SigLIP-style vision--language model
$\phi^{\mathrm{sig}}_{\mathrm{img}},\phi^{\mathrm{sig}}_{\mathrm{text}}$ and
compute cosine similarity between the image and each SP: 
\begin{equation*}
\setlength{\abovedisplayskip}{6pt}
\setlength{\belowdisplayskip}{6pt}
R_{\mathrm{glb}}(\vI^{(k)},\vp) =
\cos\!\bigl(
\phi^{\mathrm{sig}}_{\mathrm{img}}(\vI^{(k)}),
\phi^{\mathrm{sig}}_{\mathrm{text}}(\vp)
\bigr).
\end{equation*}
For the perceptual reward $R_{\mathrm{per}}$ we employ a Perception Encoder $\phi^{\mathrm{per}}_{\mathrm{img}},\phi^{\mathrm{per}}_{\mathrm{text}}$ and cosine similarity. Prompt-level scores $R_{\mathrm{glb}}(\vI^{(k)},\vp)$ and $R_{\mathrm{per}}(\vI^{(k)},\vp)$ are obtained by aggregating over SPs (uniform averaging modulated by the policy).

\noindent \textbf{Region grounding reward ($R_{\mathrm{rg}}$).}
Region-level grounding uses RegionCLIP image-region $\psi^{\mathrm{reg}}_{\mathrm{img}}(\vI,r_m)$ and text embeddings $\psi^{\mathrm{reg}}_{\mathrm{text}}(\vp)$. Given region proposals $\{r_m\}$ we compute  $s_m(\vp)$ and soft attention weights $\alpha_m(\vp)$, and define
\begin{equation*}
\setlength{\abovedisplayskip}{4pt}
\setlength{\belowdisplayskip}{4pt}
R_{\mathrm{rg}}(\vI^{(k)},\vp) = \sum_m \alpha_m(\vp)\,s_m(\vp).
\end{equation*}

This reward encourages gradients to concentrate on spatial regions that are both semantically and visually aligned with each SP, matching the behavior illustrated in \Cref{fig:overview}.

\noindent \textbf{Object consistency reward ($R_{\mathrm{oc}}$).}
For object-level localization, we use text-guided SAM2 \cite{ravi2024sam} (Florence-SAM2\footnote{\url{https://huggingface.co/spaces/SkalskiP/florence-sam/blob/main/checkpoints/sam2_hiera_large.pt}}) to obtain soft masks $\{\vM_j\}$ and confidences $\{a_j\}$ for each semantic primitive. For each SP $p$, we query SAM2 with the text description and optional point prompts derived from its coarse localization (\eg from the region-level gradients), yielding soft foreground masks $M_j$ and their confidences $a_j$. Mixture weights $\omega_j$ are formed via a softmax over $a_j$. We compute an object alignment score $F_{\mathrm{obj}}(M_j,g_{\mathrm{SP}}(\vp))$ that rewards correct semantics in the mask and penalizes leakage in the background. The object reward for SP $p$ is
\begin{equation*}
\setlength{\abovedisplayskip}{4pt}
\setlength{\belowdisplayskip}{4pt}
R_{\mathrm{oc}}(\vI^{(k)},\vp)
=
\sum_j \omega_j
\,F_{\mathrm{obj}}(M_j,g_{\mathrm{SP}}(\vp)),
\end{equation*}
and is further modulated by the add/remove intent scalar $s_{\mathrm{obj}}\in[-1,1]$ predicted by the adaptive policy.

\noindent \textbf{Human Preference Reward ($R_{\mathrm{hps}}$).}
For $R_{\mathrm{hps}}$, we use HPSv2, a pretrained human preference scorer that takes $(\vI^{(k)},\vp)$ as input and outputs a scalar score. We normalize this score with a fixed
running mean and variance so that it is numerically comparable to the other rewards and can be combined without further scaling:
\[
R_{\mathrm{hps}}(\vI^{(k)},\vp)
=
\mathrm{norm}\bigl(
H_{\mathrm{HPS}}(\vI^{(k)},\vp)
\bigr).
\]
In practice, $R_{\mathrm{hps}}$ is evaluated on the full prompt and primarily stabilizes overall aesthetic quality and prompt adherence.

\noindent \textbf{VQA reward ($R_{\mathrm{vqa}}$).}
For $R_{\mathrm{vqa}}$ we use the Qwen-2.5-VL 3B model,
accessed via the HuggingFace Transformers interface. 
For each Q\&A pair $(q,a^\star)$ produced by ChatGPT, we feed $(\vI^{(k)},\vp)$ into Qwen-2.5-VL and obtain the token-level logits $\{\ell_t\}_{t=1}^{T^\star}$ for the answer sequence $a^\star\!-\!(a_t^\star)_{t=1}^{T^\star}$. 
We then form the VQA reward from these logits.
In practice, we cap $T^\star$ to a reasonable answer length (\eg $T^\star\leq 70$ tokens).

\noindent \textbf{Cosine similarity backbone.}
All rewards except $R_{\mathrm{hps}}$ and $R_{\mathrm{vqa}}$ are
implemented as cosine similarities between the embeddings
of semantic primitives and the current image at step $k$.
Gradients are obtained via automatic differentiation through
the corresponding vision--language encoders.

\noindent \textbf{KL Tether for Image Editing.}
For all image editing experiments, we enable the clean-latent KL tether $g_{\mathrm{KL},k}$ from Eq.~(5). The tether is computed in the clean latent space $\tilde{\vz}^{(k)}$ and back-propagated through $\mathrm{Den}_\theta$ using its Jacobian $J_{\mathrm{Den}}$. We keep the KL strength
$\lambda_{\mathrm{KL}}\!=\!1.5$ fixed across steps and applied only when a source image is provided. For pure text-to-image
generation, we set $\lambda_{\mathrm{KL}}\!=\!0$ so that the sampler targets the prompt-tilted distribution without anchoring to a
particular source latent.

\begin{figure}[t!]
    \centering
    \includegraphics[width=0.85\linewidth]{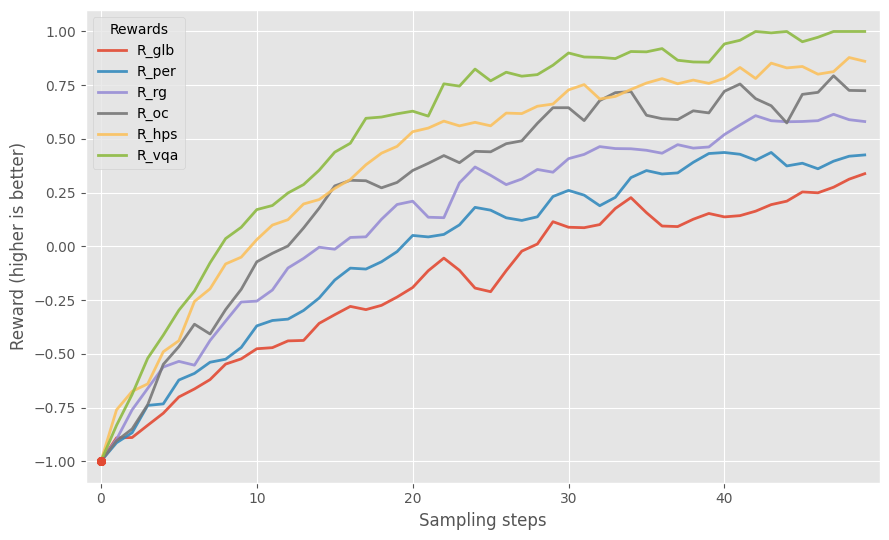}
    \vspace{-0.3cm}
    \caption{
    \textbf{Reward progression over time.}
    }
    \label{fig:rewardplots}
    \vspace{-0.2cm}
\end{figure}

\noindent \textbf{Editing vs.\ Generation Configurations.} For text-to-image experiments, we use all rewards except the object-consistency reward $R_{\mathrm{oc}}$, which is less
relevant in the absence of a reference layout. For image editing, we enable the full set of rewards
$\{R_{\mathrm{glb}},R_{\mathrm{per}},R_{\mathrm{oc}},R_{\mathrm{rg}}, R_{\mathrm{hps}},R_{\mathrm{vqa}}\}$, dynamic reward weighting, reward-aware step sizes, and the KL tether. 
As shown in \Cref{fig:rewardplots}, all reward components in \modelnamenc{} exhibit consistent and stable improvement over the course of sampling. Starting from an initial value of $-1$, the global semantics reward $R_{\text{glb}}$, perceptual reward $R_{\text{per}}$, region grounding reward $R_{\text{rg}}$, object consistency reward $R_{\text{oc}}$, human-preference reward $R_{\text{hps}}$, and VQA reward $R_{\text{vqa}}$ all trend upward with natural fluctuations, eventually converging to high positive values. The smooth yet spiky trajectories indicate that the system is actively exploring while steadily refining the sample quality under each objective, rather than overfitting to any single reward. Taken together, these qualitative dynamics demonstrate that \modelnamenc{} effectively coordinates and optimizes all reward signals, confirming that the full reward pipeline operates as intended.

\section{Additional Results}
\noindent \textbf{Text-to-Image Generation.}
We perform additional text-to-image generation evaluation on \textsc{GenEval}.
As shown in \Cref{tab:compositionality}, \modelnamenc{} consistently improves compositional faithfulness over both backbone models and the ReNO baseline. Starting from weaker backbones such as PixArt-$\alpha$ DMD and Flux, \modelnamenc{} lifts the mean score from 0.45→0.65 and 0.64→0.81, respectively, and further improves over ReNO by +0.06 and +0.09 in overall performance. The gains are largest on the most compositional sub-tasks: for PixArt-$\alpha$, Two objects and Counting increase from 0.38/0.46 to 0.77/0.65, and for Flux from 0.80/0.64 to 0.97/0.90. Even on the strong Qwen backbone, \modelnamenc{} improves the overall performance from 0.83 to 0.91 and surpasses ReNO on all metrics, notably boosting Position from 0.27→0.47 and Color Attribution from 0.71→0.84. As a result, Qwen + \modelnamenc{} achieves the best overall \textsc{GenEval} performance, outperforming powerful off-the-shelf models such as SDXL, DALL-E 3, and SD3 (8B), whose mean scores remain in the 0.55–0.68 range.\looseness-1

These quantitative gains stem from the way \modelnamenc{} integrates diverse, task-aligned rewards into test-time optimization. Instead of relying primarily on a global alignment signal as in ReNO, \modelnamenc{} evaluates a heterogeneous set of differentiable rewards covering semantic and perceptual alignment, regional and object-level consistency, and QA-style reasoning and fuses their gradients through a prompt-aware adaptive policy that adjusts reward weights and step sizes along the denoising trajectory. This richer, spatially and semantically grounded feedback allows the sampler to correct fine-grained failures such as incorrect counts, swapped colors, or mislocalized objects, while preserving the overall realism of the backbone generator. Consequently, \modelnamenc{} is better able to satisfy complex multi-object, attribute, and localization constraints, which is reflected in its strong improvements on Two objects, Counting, Position, and Color Attribution compared to both unmodified backbones and prior reward-guided baselines.
\begin{table}[t]
\centering
\footnotesize
\setlength{\tabcolsep}{3pt}
\caption{\textbf{T2I generation on \textsc{GenEval}}.}
\label{tab:compositionality}
\vspace{-0.3cm}
\resizebox{0.99\columnwidth}{!}{%
\begin{tabular}{lccccccc}
\toprule
\textbf{Model} & \textbf{Overall} $\uparrow$ & \textbf{Single} $\uparrow$ & \textbf{Two} $\uparrow$ & \textbf{Counting} $\uparrow$ & \textbf{Colors} $\uparrow$ & \textbf{Position} $\uparrow$ & \textbf{Color Attribution} $\uparrow$ \\
\midrule
SD v2.1                    & 0.50 & 0.98 & 0.51 & 0.44 & 0.85 & 0.07 & 0.17 \\
SDXL                       & 0.55 & 0.98 & 0.74 & 0.39 & 0.85 & 0.15 & 0.23 \\
IF-XL                      & 0.61 & 0.97 & 0.74 & 0.66 & 0.81 & 0.13 & 0.35 \\
PixArt-$\alpha$            & 0.48 & 0.98 & 0.50 & 0.44 & 0.80 & 0.08 & 0.07 \\
DALL-E 2                   & 0.52 & 0.94 & 0.66 & 0.49 & 0.77 & 0.10 & 0.19 \\
DALL-E 3                   & 0.67 & 0.96 & 0.87 & 0.47 & 0.83 & 0.43 & 0.45 \\
SD3 (8B)                   & 0.68 & 0.98 & 0.84 & 0.66 & 0.74 & 0.40 & 0.43 \\
\midrule
(1) PixArt-$\alpha$ DMD    & 0.45 & 0.95 & 0.38 & 0.46 & 0.76 & 0.05 & 0.09 \\
\rowcolor{gray!10} (1) + ReNO                & 0.59 & 0.98 & 0.72 & 0.58 & 0.85 & 0.15 & 0.27 \\
\rowcolor{rewardpurple!10} \textbf{(1) + \modelnamecolor}    & 0.65 & 0.99 & 0.77 & 0.65 & 0.89 & 0.21 & 0.33 \\
\midrule
(2) Flux           & 0.64 & 0.98 & 0.80 & 0.64 & 0.78 & 0.18 & 0.43 \\
\rowcolor{gray!10} (2) + ReNO             & 0.72 & 0.99 & 0.90 & 0.79 & 0.87 & 0.21 & 0.56 \\
\rowcolor{rewardpurple!10} \textbf{(2) + \modelnamecolor}    & 0.81 & 0.99 & 0.97 & 0.90 & 0.95 & 0.39 & 0.72 \\
\midrule
(5) Qwen                   & 0.83 & 0.99 & 0.98 & 0.92 & 0.92 & 0.27 & 0.71 \\
\rowcolor{gray!10} (5) + ReNO             & 0.85 & 0.99 & 0.98 & 0.94 & 0.95 & 0.35 & 0.75 \\
\rowcolor{rewardpurple!10} \textbf{(5) + \modelnamecolor}  & 0.91 & 0.99 & 0.99 & 0.97 & 0.98 & 0.47 & 0.84 \\
\bottomrule
\end{tabular}%
}
\vspace{-0.2cm}
\end{table}

\begin{figure*}[t!]
  \centering
  \includegraphics[width=0.95\textwidth]{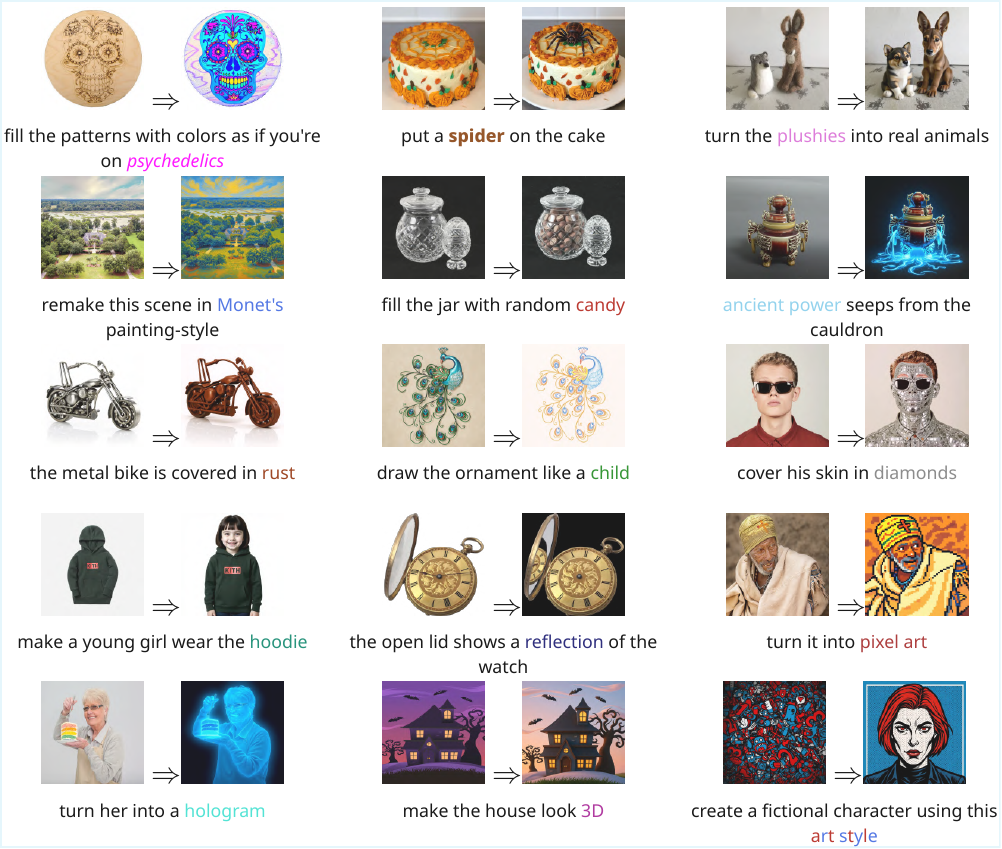}
  \vspace{-0.4cm}
    \caption{\textbf{Image Editing Qualitative Results with Flux + \modelnamenc{}}. For each input image on the left, \modelnamenc{} is instructed to apply a targeted edit (text below), and the right image shows the generated result. Tasks span from global scene modifications and object-level edits to very fine-grained, localized edits.}
  \label{fig:editsupp1}
    \vspace{-0.2cm}
\end{figure*}

\begin{figure*}[t!]
  \centering
  \includegraphics[width=0.95\textwidth]{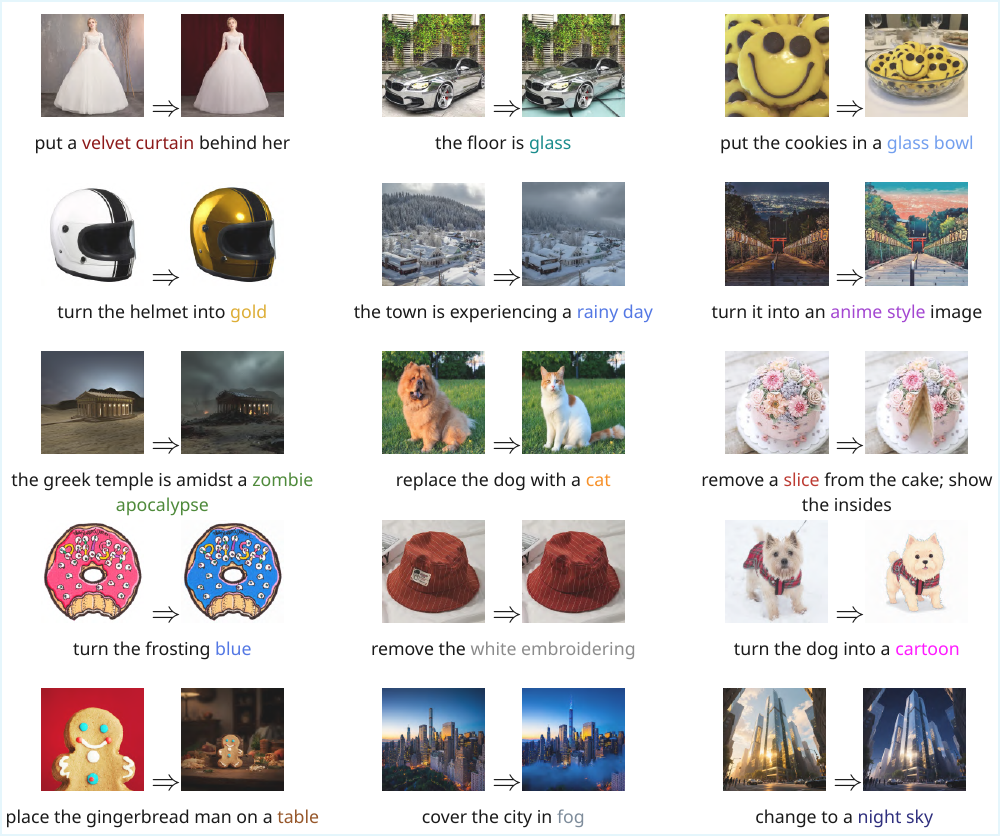}
  \vspace{-0.4cm}
  \caption{\textbf{Image Editing Qualitative Results with Qwen Image + \modelnamenc{}}. For each input image on the left, \modelnamenc{} is instructed to apply a targeted edit (text below), and the right image shows the generated result. Tasks span from global scene modifications and object-level edits to very fine-grained, localized edits.}
  \label{fig:editsupp2}
    \vspace{-0.3cm}
\end{figure*}

\begin{figure*}[t!]
  \centering
  \includegraphics[width=0.99\textwidth]{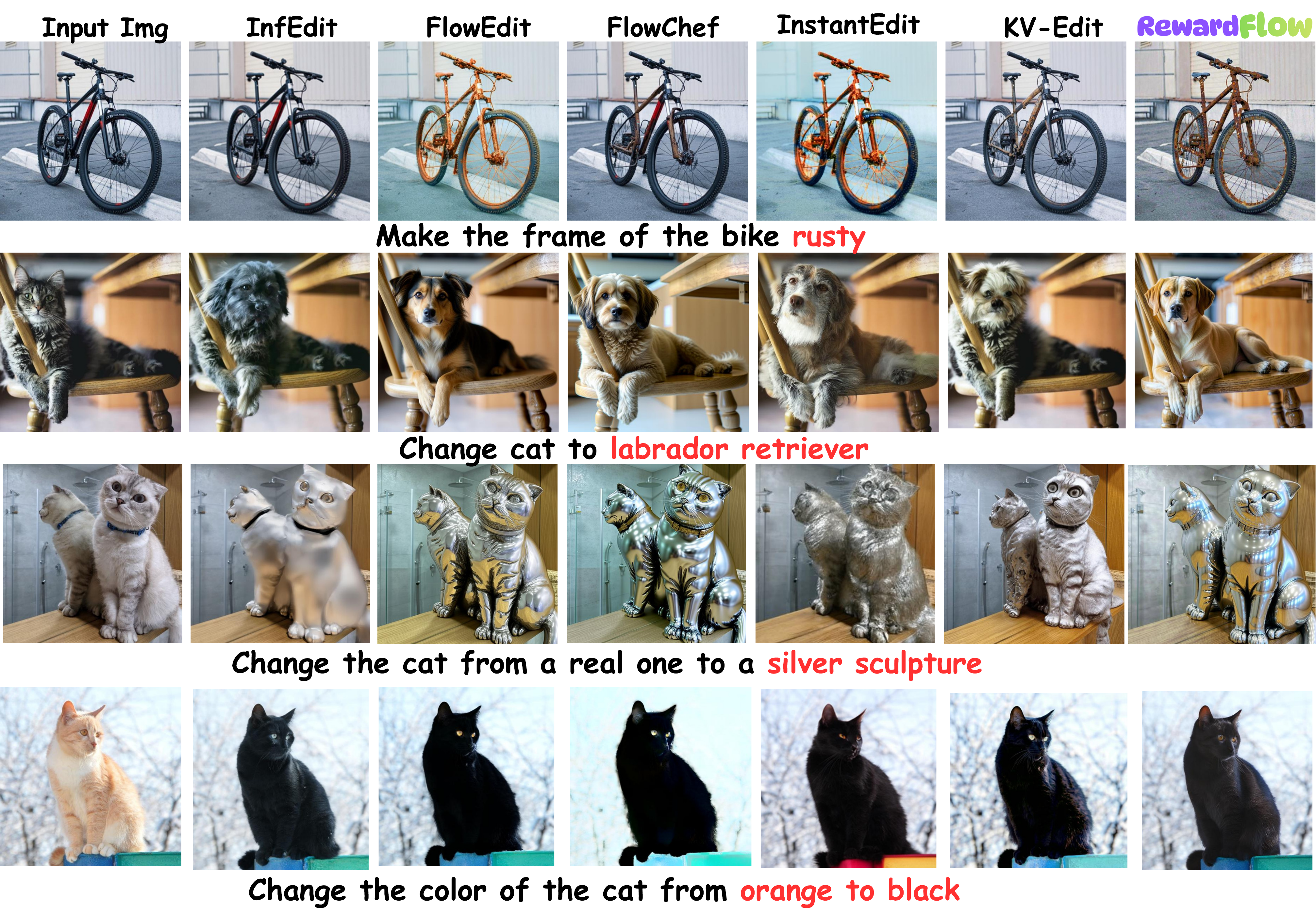}
    \vspace{-0.4cm}
  \caption{\textbf{Image Editing Qualitative Comparisons.} Comparison across a range of challenging edits, such as diverse attribute-, style-, and object-level transformations. Each row shows the source image followed by results from strong baselines and \modelnamenc{}.} 
  \label{fig:fullpage_image3}
  \vspace{-0.2cm}
\end{figure*}

\begin{table}[t]
\centering
\caption{\textbf{VLM Comparison on \textsc{PIE-Bench}.}}
\vspace{-0.3cm}
\label{tab:vlms}
\setlength{\tabcolsep}{4pt}
\renewcommand{\arraystretch}{0.95}
\resizebox{0.95\linewidth}{!}{
\begin{tabular}{lccccccc}
\toprule
\textbf{VLMs} &
\makecell{\textbf{Distance} $\downarrow$\\($\times10^3$)} &
\textbf{PSNR} $\uparrow$ &
\makecell{\textbf{LPIPS} $\downarrow$\\($\times10^3$)} &
\makecell{\textbf{MSE} $\downarrow$\\($\times10^4$)} &
\makecell{\textbf{SSIM} $\uparrow$\\($\times10^2$)} &
\textbf{Whole} $\uparrow$ &
\textbf{Edited} $\uparrow$ \\
\midrule
\rowcolor{rewardpurple!20} \textbf{Qwen 2.5VL 3B$\dagger$} & \textbf{7.64} & \textbf{32.09} & \textbf{38.47} & \textbf{33.57} & \textbf{90.21} & \textbf{29.78} & \textbf{27.57} \\
LLaMa-4-8B
& 6.57 & 33.43 & 37.19 & 31.31 & 91.33 & 30.44 & 28.82 \\
Qwen 3 Next-34B
& 6.53 & 32.34 & 38.05 & 32.76 & 91.49 & 31.01 & 29.03 \\
\bottomrule
\end{tabular}
}
\vspace{-0.4cm}
\end{table}

\noindent \textbf{Ablation on VLMs.}
We further conduct an ablation study by replacing the visual-language model (VLM) used for $R_{\text{vqa}}$ with different architectures.  As shown in Table~\ref{tab:vlms}, the overall performance remains relatively stable when scaling from 3B to 8B parameters, indicating that moderate model scaling yields limited benefit for this task.  However, substituting with the larger and more recent {Qwen3-Next-34B} model leads to a noticeable $\sim$7\% improvement across most evaluation metrics, suggesting that more capable VLMs enhance semantic reasoning in the reward estimation process, however, at the expense of increased computational overhead.\looseness-1

\section{Additional Qualitative Results}
\noindent \textbf{Image Editing Qualitative Results.}
Using Flux as the base model, as shown in \Cref{fig:editsupp1}, \modelnamenc{} follows a wide variety of fine-grained instructions while preserving background layout and image identity. \modelnamenc{} can perform strong stylistic changes, such as recoloring the carved wooden ornament “as if on psychedelics,” translating a natural landscape into Monet’s painting style, and turning a portrait into pixel art, all while keeping shapes and composition intact. Our proposed method also accurately handles object insertion and modification: a spider is added on top of the cake, plush toys are turned into realistic animals, a jar is filled with random candy, and “ancient power” is made to seep from the cauldron with coherent lighting. Local attribute edits are also precisely localized, \eg metal parts of the bike are rusted without any corruptions, the ornament is redrawn in a child-like manner, the subject’s skin is covered with diamonds, and the pocket watch lid reflects the watch face without hallucinating unrelated content. Finally, \modelnamenc{} successfully performs more abstract edits such as making a young girl wear the same hoodie, turning the woman into a hologram, making the cartoon house appear 3D, and synthesizing a new fictional character inspired by a textured input image. Across all examples, edits are restricted to instruction-relevant regions and avoid semantic leakage into the rest of the scene.\looseness-1

With Qwen Image as the backbone, shown in \Cref{fig:editsupp2}, \modelnamenc{} exhibits similarly precise and diverse editing capabilities. Global scene edits include converting a sunny town into a rainy day, covering a city with fog, and changing a bright skyline to a night sky, while preserving camera pose and urban geometry. Attribute and material changes are handled cleanly, \eg a velvet curtain is placed behind the bride, the showroom floor becomes glass, helmet material is changed to gold, and cake frosting is recolored blue without affecting decorations. \modelnamenc{} also supports challenging object-level manipulations, such as putting cookies into a glass bowl, turning a dog into a cartoon, and placing the gingerbread man onto a table with consistent perspective. Fine, localized modifications, such as removing the embroidered text on the hat, removing a slice from the cake and revealing the inside, and staging a “zombie apocalypse” around a Greek temple, are executed while maintaining sharp structure and coherent lighting. Results demonstrate \modelnamenc{} generalizes across backbones and instruction types, delivering semantically faithful, spatially localized edits from global scene transforms down to pixel-level adjustments.

\Cref{fig:fullpage_image3} presents a qualitative comparison between \modelnamenc{} and recent image editing methods, including InfEdit, FlowEdit, FlowChef, InstantEdit, and KV-Edit, under the same input image and text instruction. The figure covers a range of challenging edit types, including material transformation, object-level semantic replacement, and color editing.
In the first row, the instruction asks to make the frame of the bike rusty. Baseline methods exhibit different failure modes, \eg some methods under-edit the image and leave large parts of the bicycle frame nearly unchanged (such as InfEdit, FlowChef, and KV-Edit), while others apply the rusty texture too aggressively or inconsistently, affecting broader regions and introducing unnatural appearance changes (such as FlowEdit and InstantEdit). In contrast, \modelnamenc{} successfully transfers the rusty material appearance onto the bicycle frame while preserving the overall structure, viewpoint, wheel geometry, and background scene, resulting in a more coherent and realistic edit.
In the cat-to-labrador transformation (second row), several baselines either fail to fully realize the target dog breed or generate inconsistent appearances, whereas \modelnamenc{} produces a more convincing labrador retriever that remains in the same position on the chair, while keeping the surrounding environment intact.
For the real-cat to silver-sculpture edit, baselines either fail to fully impose the metallic sculptural material or introduce artifacts in shape and surface reflectance, whereas \modelnamenc{} renders metallic texture and reflective highlights, while preserving the original pose, object boundaries, and scene composition.
Finally, in the fourth row, baselines sometimes over-darken the image, alter contrast unnaturally, or fail to perform a clean color transition, whereas \modelnamenc{} produces a cleaner black cat while maintaining the cat's silhouette, eye color, and overall scene context.

\begin{figure}[t!]
    \centering
    \includegraphics[width=0.99\linewidth]{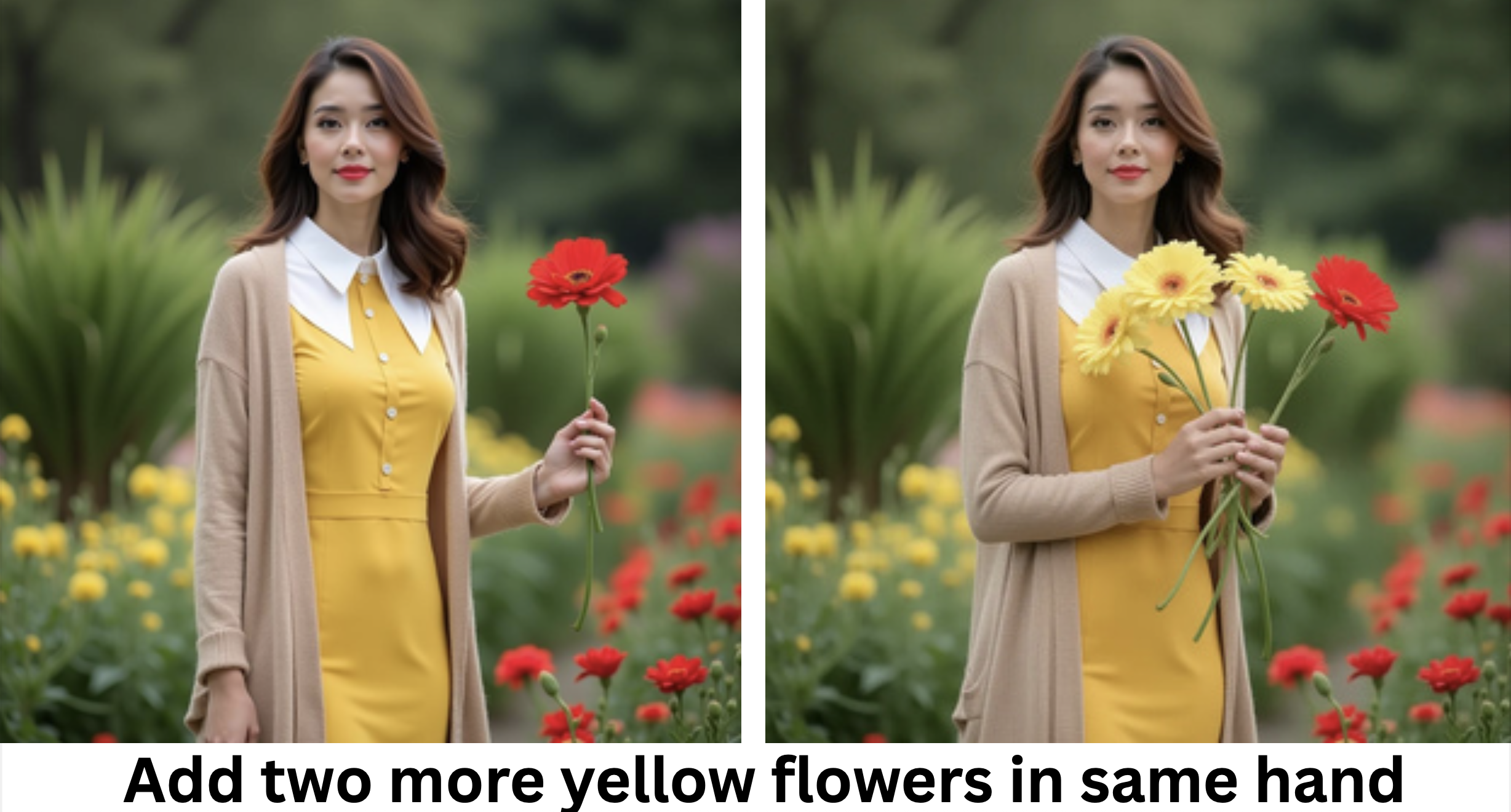}
    \vspace{-0.2cm}
    \caption{\textbf{\modelnamenc{} counting failure case.}}
    \label{fig:rewardfail}
    \vspace{-0.2cm}
\end{figure}

\noindent \textbf{Failure Modes.}
While robust, RewardFlow is bounded by its components. A primary failure mode arises from {VQA limitations} in fine-grained reasoning like counting. As shown in Figure.~\ref{fig:rewardfail}, if VQA model fails to accurately count small objects, the reward signal becomes uninformative.

\begin{figure*}[t!]
  \centering
  \includegraphics[height=.99\textheight,keepaspectratio]{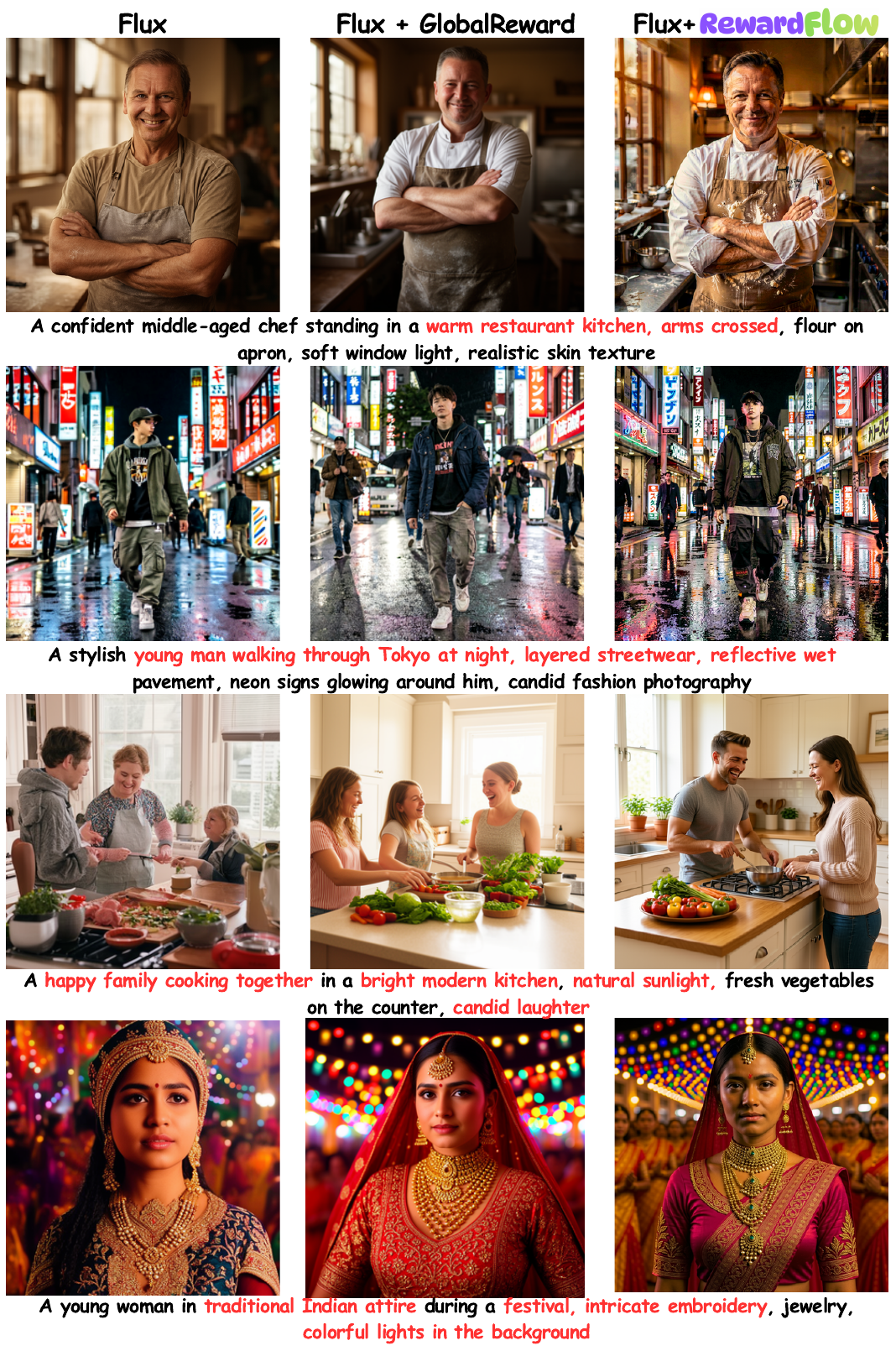}
    \vspace{-0.3cm}
  \caption{\textbf{Text-to-image qualitative results with Flux as backbone.} Qualitative comparison of Flux, Flux + Global Reward, and Flux + \modelnamenc{} across diverse prompts.}
  \label{fig:fullpage_image1}
\end{figure*}

\noindent \textbf{Text-to-Image Generation Qualitative Results.}
\Cref{fig:fullpage_image1} presents qualitative comparisons for text-to-image generation with the Flux backbone under three inference settings: vanilla Flux, Flux guided by a global matching reward ({Flux + GlobalReward}), and the full reward-augmented model, \modelnamenc{} ({Flux + RewardFlow}). Across a diverse set of prompts, including a chef portrait in a restaurant kitchen, a street-fashion scene in nighttime Tokyo, a multi-person family cooking scene, and a culturally specific festival portrait, the vanilla backbone generally captures the coarse scene semantics but frequently under-specifies fine-grained attributes, weakens environmental grounding, and exhibits limited compositional precision. Incorporating only the global reward improves overall prompt alignment and image aesthetics, yet the generations still miss localized details and precise relational cues, particularly in clothing structure, scene context, object placement, and human interaction.
In contrast, \modelnamenc{} consistently produces samples with stronger semantic fidelity, improved spatial and contextual grounding, and higher perceptual coherence.
In the chef example, \modelnamenc{} better realizes the warm kitchen environment, apron texture, flour details, and realistic skin appearance. In the Tokyo street scene, \modelnamenc{} more faithfully captures the wet-pavement reflections, while in the family cooking example, \modelnamenc{} yields more natural multi-person interaction, and better localized food and countertop details. In the festival portrait, \modelnamenc{} more accurately renders traditional Indian attire through richer embroidery, more convincing jewelry, and a stronger festive lighting atmosphere.

\begin{algorithm}[t]
\caption{\modelnamenc{}: Prompt-aware multi-reward Langevin editing}
\label{alg:main}
\begin{algorithmic}[1]
\STATE \textbf{Input:} image $\vx$, prompt $\vp$, steps $K$
\STATE $\vz_0 \leftarrow \Enc(\vx)$
\STATE $\{\vp_m\}_{m=1}^M \leftarrow \textsc{ExtractSPs}(\vp)$\COMMENT{semantic primitives}
\STATE $(q,a^\star) \leftarrow \textsc{MakeQA}(\vx,\vp)$ \COMMENT{fixed once}
\STATE Initialize running stats $\{\mu_i,\sigma_i\}_{i}$ for all heads
\STATE Sample $\varepsilon \sim \mathcal{N}(0,\vI)$,
       set $\vz^{(0)} \leftarrow \alpha_{\bar t} \vz_0 + \sigma_{\bar t}\varepsilon$,
       $t_0 \leftarrow \bar t$
\FOR{$k = 0$ \TO $K-1$}
  \STATE $\tilde \vz^{(k)} \leftarrow \mathrm{Den}(\vz^{(k)}, t_k, \vp)$
  \STATE $\vI^{(k)} \leftarrow \Dec(\tilde \vz^{(k)})$
  \STATE \COMMENT{SP-wise rewards: compute over $\vp \in \mathrm{SP}(\vp)$}
  \STATE \textbf{Initialize} score vectors: $\mathbf{v}_{\text{glb}}, \mathbf{v}_{\text{per}}, \mathbf{v}_{\text{rg}}, \mathbf{v}_{\text{oc}} \leftarrow [\,]$
\FOR{each $\vp \in \mathrm{SP}(\vp)$}
  \STATE $\mathbf{v}_{\text{glb}}.\text{append}(\textsc{GlobReward}(\vI^{(k)}, \vp))$
  \STATE $\mathbf{v}_{\text{per}}.\text{append}(\textsc{PercReward}(\vI^{(k)}, \vp))$
  \STATE $\mathbf{v}_{\text{rg}}.\text{append}(\textsc{RegionReward}(\vI^{(k)}, \vp))$
  \STATE $\mathbf{v}_{\text{oc}}.\text{append}(\textsc{ObjReward}(\vI^{(k)}, \vp))$
\ENDFOR

\STATE \COMMENT{Adaptive fusion of SP scores per head}
\FOR{$h \in \{\text{glb, per, rg, oc}\}$}
  \STATE $\boldsymbol{\ell}_h \leftarrow \textsc{ComputeWeights}(\mathbf{v}_h, t_k, \mathrm{SP}(\vp))$
  \STATE $\boldsymbol{\omega}_h \leftarrow \operatorname{softmax}(\boldsymbol{\ell}_h)$
  \STATE $R_h^{(k)} \leftarrow \sum_{j} \boldsymbol{\omega}_h[j] \cdot \mathbf{v}_h[j]$ \COMMENT{weighted fusion}
\ENDFOR

  \STATE \COMMENT{Prompt-wise rewards: computed once per step}
  \STATE $R_{\text{hps}}^{(k)} \leftarrow \textsc{HPSReward}(\vI^{(k)}, \vp)$
  \STATE $R_{\text{vqa}}^{(k)} \leftarrow \textsc{VQAReward}(\vI^{(k)}, q, a^\star)$
  \STATE \COMMENT{Normalize each head}
  \FOR{each head $i$}
    \STATE Update $\mu_i,\sigma_i$; \ \ $\bar R_i^{(k)} \leftarrow (R_i^{(k)}-\mu_i)/(\sigma_i+\epsilon)$
  \ENDFOR
  \STATE \COMMENT{Prompt-aware adaptive weights (Sec.~\ref{subsec:adaptive})}
  \STATE $\{w_i^{(k)}\}_i \leftarrow \textsc{ComputeWeights}\!\big(\{\bar R_i^{(k)}\}, t_k, \vp\big)$
  \STATE $R_{\text{tot}}^{(k)} \leftarrow \sum_i w_i^{(k)} \bar R_i^{(k)}$
  \STATE \COMMENT{Fused reward gradient in latent space  (Sec.~\ref{subsec:prelim-langevin})}
  \STATE $g_\vI^{(k)} \leftarrow \sum_i w_i^{(k)} \nabla_{\vI^{(k)}} \bar R_i^{(k)}$
  \STATE $g_{R_{\text{tot}},k} \leftarrow \lambda_R
         J_{\mathrm{Den}}(\vz^{(k)},t_k,\vp)^\top
         J_{\mathrm{Dec}}(\tilde \vz^{(k)})^\top g_\vI^{(k)}$
  \STATE \COMMENT{Backbone drift and KL tether (Sec.~\ref{subsec:kltether})}
  \STATE $f_k \leftarrow v_\theta(\vz^{(k)}, t_k, \vp)$
  \STATE $g_{\text{KL},k} \leftarrow -\,\lambda_{\text{KL}}\,
         J_{\mathrm{Den}}(\vz^{(k)},t_k,\vp)^\top(\tilde \vz^{(k)} - \vz_0)$
  \STATE \COMMENT{Reward-aware step size (Sec.~\ref{subsec:adaptive})}
  \STATE $\eta_k \leftarrow \textsc{StepSize}(R_{\text{tot}}^{(k)})$; \ $\xi_k \sim \mathcal{N}(0,\vI)$
  \STATE \COMMENT{Langevin update}
  \STATE $\vz^{(k+1)}\leftarrow\vz^{(k)}
         + \eta_k (f_k + g_{R_{\text{tot}},k} + g_{\text{KL},k})$
         
         \qquad\qquad\qquad\ $+ \sqrt{2\gamma(t_k)\eta_k}\xi_k$
  \STATE $t_{k+1} \leftarrow t_k - \eta_k$
\ENDFOR
\STATE \textbf{return} $\hat \vI \leftarrow \Dec\big(\mathrm{Den}_\theta(\vz^{(K)},0,\vp)\big)$
\end{algorithmic}
\end{algorithm}

\end{document}